\documentclass[11pt]{article}

\usepackage[preprint]{acl}

\usepackage{times}
\usepackage{latexsym}

\usepackage[T1]{fontenc}

\usepackage[utf8]{inputenc}

\usepackage{microtype}

\usepackage{inconsolata}

\usepackage{graphicx}

\usepackage[table]{xcolor}
\usepackage{graphicx}
\usepackage{siunitx}
\usepackage{amsthm,amsfonts,amsmath}
\usepackage{array,multirow}
\usepackage{booktabs}
\usepackage{listings}
\usepackage{float}
\usepackage{bm}
\floatname{listing}{Listing}
\newfloat{listing}{thp}{lop}
\usepackage{subcaption}
\usepackage{tcolorbox}

\newcommand{\blue}[1]{\textcolor{blue}{#1}}

\title{Query Symbolically or Retrieve Semantically? \\A Dataset and Method for Semi-Structured Question Answering}

\author{
 \textbf{Mateusz Czy\.{z}nikiewicz\textsuperscript{1}},
 \textbf{Ryszard Tuora\textsuperscript{1}},
 \textbf{Adam Kozakiewicz\textsuperscript{1}},
\\
 \textbf{Tomasz Zi\k{e}tkiewicz\textsuperscript{1}},
 \textbf{Mateusz Gali\'{n}ski\textsuperscript{1}},
 \textbf{Micha\l{} T. Godziszewski\textsuperscript{1}},
\\
 \textbf{Micha\l{} Karpowicz\textsuperscript{1}},
 \textbf{Timothy Hospedales\textsuperscript{2}},
 \textbf{Cristina Cornelio\textsuperscript{2}}
\\
\\
 \textsuperscript{1}Samsung AI Warsaw,
 \textsuperscript{2}Samsung AI Cambridge,
\\
 \small{
   \textbf{Correspondence:} \href{mailto:c.cornelio@samsung.com}{c.cornelio@samsung.com}
 }
}

\begin{document}

\maketitle

\begin{abstract}

Retrieval-Augmented Generation (RAG) systems for question answering typically retrieve evidence by semantic similarity between the query and document chunks. 
While effective for unstructured text, this approach is less reliable on semi-structured corpora where answering may require exact filtering, aggregation, or exhaustive retrieval over structured attributes 
across multiple documents. 
Symbolic approaches support such operations, but they are often brittle on noisy natural-language corpora.
We address this gap with \textbf{DualGraph}, a RAG framework that represents 
documents through two complementary views: a \textit{Textual Knowledge Graph} for semantic retrieval and a \textit{Symbolic Knowledge Graph} for symbolic querying over typed subject--predicate--object triples. Building on these two components, we provide multiple strategies for selecting or combining semantic and symbolic evidence.
We also introduce \textbf{SpecsQA}, a benchmark from a commercial shopping website with semi-structured product documents and manually curated questions spanning open-ended and specification-oriented retrieval. Experiments show that DualGraph consistently outperforms state-of-the-art dense-retrieval, GraphRAG, symbolic, and table-oriented baselines across question types.
Code and data are available at \url{https://github.com/corneliocristina/DualGraphRAG}.

\end{abstract}

\section{Introduction}
Retrieval-Augmented Generation (RAG) has become a standard approach for question answering by grounding language models on external evidence retrieved from large document collections. Most existing RAG systems \cite{2020_RAG_paper} retrieve evidence according to semantic similarity between the query and document chunks in an embedding space. This works well when relevant information is expressed in localized natural-language (NL) passages, but becomes less reliable for semi-structured corpora that combine free-form text with structured specifications, tables, or attributes distributed across multiple documents.

Many real-world questions over such corpora require operations beyond what semantic similarity alone can provide. Queries such as \textit{Which Samsung phones support both wireless charging and eSIM?} or \textit{List all TVs with HDMI 2.1 and refresh rates above 120Hz} require exact filtering, aggregation, or exhaustive listing over structured attributes. 
While recent graph-based RAG systems \cite{gutierrez2025hipporag2,zhuang2025linearraglineargraphretrieval} improve multi-hop and cross-document retrieval, they still lack formal guarantees for precise operations such as exhaustive enumeration or arithmetic aggregation. Symbolic approaches \cite{chepurova2025wikontic} can support exact querying, but they are often not robust to noisy natural-language questions and heterogeneous document structures.

This gap appears in many real-world settings, including enterprise documentation, local-file assistants, and customer-support systems over product catalogs or manuals, where users ask both open-ended questions and precise specification-oriented queries.
To address this limitation, we introduce \textbf{DualGraph}, a retrieval framework for question answering over semi-structured corpora. DualGraph represents the same documents collection through two complementary graph views: a \textbf{Textual Knowledge Graph} (TKG), optimized for semantic retrieval over natural-language descriptions, and a \textbf{Symbolic Knowledge Graph} (SKG), designed for structured querying over normalized subject--predicate--object triples. Our framework supports multiple retrieval strategies that combine semantic and symbolic evidence in different ways, including direct retrieval over either graph, fallback mechanisms, concatenation of retrieved contexts, and query-dependent routing.

We also introduce \textbf{SpecsQA}, a benchmark built from a time-specific snapshot of the Samsung UK website. The benchmark contains semi-structured product documents combining natural-language descriptions with specification tables. Existing QA benchmarks \cite{yang-etal-2018-hotpotqa,trivedi-etal-2022-musique} rarely combine structured and unstructured information, and mixed text-table datasets typically focus on questions grounded within a single document or table instance \cite{strich-etal-2026-t2}. In contrast, SpecsQA evaluates corpus-level retrieval and reasoning, requiring systems to identify and combine evidence across multiple documents and content types before answering. Because the benchmark is built from a website snapshot that evolves over time, it also reduces the likelihood that questions can be answered through memorization by pre-trained language models alone.

Experiments show that DualGraph consistently outperforms state-of-the-art dense-retrieval, graph-based, symbolic, and table-oriented baselines across evaluation metrics. The improvement is especially pronounced on questions like identifying all products satisfying a set of constraints, a setting where current RAG systems are particularly limited. These results show that the textual and symbolic graph views provide complementary strengths: the SKG improves precise filtering and exhaustive list retrieval, while the TKG provides robustness for more open-ended or underspecified questions.

To summarize, our contributions are:
\textbf{1)} We introduce \textbf{DualGraph}, a RAG architecture that combines semantic and symbolic retrieval through complementary graph representations;
\textbf{2)} We release \textbf{SpecsQA}, a benchmark for question answering over semi-structured corpora with both natural-language and symbolic answers; 
\textbf{3)} We empirically show that current RAG systems remain limited on semi-structured QA, and that combining semantic and symbolic retrieval improves performance.

\subsection{Related Work} 

Retrieval-Augmented Generation (RAG) \cite{2020_RAG_paper,gao2023retrieval} augments language models with external evidence retrieved at inference time. Standard index-based RAG systems retrieve text chunks according to semantic similarity to the user query and have shown strong performance on open-domain question answering tasks. However, semantic retrieval becomes less reliable in large-scale semi-structured corpora, especially when answering requires exact filtering, aggregation, or exhaustive retrieval over structured attributes distributed across multiple documents.

To improve retrieval over complex corpora, recent GraphRAG approaches \cite{edge2025localglobalgraphrag,peng2025graphragSurvey,zhuang2025linearraglineargraphretrieval,hu2025graggraphretrievalaugmentedgeneration,gutierrez2025hipporag2,DBLP:conf/iclr/SarthiATKGM24} augment retrieval with graph representations that capture entities, relations, and cross-document structure. These methods improve multi-hop and global retrieval, but they do not provide formal guarantees for operations such as exhaustive listing or exact filtering. Complementary symbolic and logic-based systems \cite{chepurova2025wikontic,mo2025kggen} support exact inference, but typically require structured inputs and are less robust to noisy natural-language text. More recent agentic RAG systems \cite{agenticRAGSurvey2026,a-rag2026,arag2025,ragenta2025,singh2025agentic} further extend retrieval pipelines through iterative exploration and reasoning steps.

Question answering benchmarks have traditionally focused either on unstructured text or on structured information such as tables and knowledge graphs. Popular text-only datasets include NaturalQuestions \cite{kwiatkowski-etal-2019-natural}, TriviaQA \cite{joshi-etal-2017-triviaqa}, HotpotQA \cite{yang-etal-2018-hotpotqa}, and MuSiQue \cite{trivedi-etal-2022-musique}. Structured and semi-structured QA benchmarks include knowledge-graph datasets such as WebQuestions \cite{berant-etal-2013-semantic}, ComplexWebQuestions \cite{talmor-berant-2018-web}, 2WikiMultiHopQA \cite{2WikiMultiHopQA}, and BMW-KG \cite{ragonite}, as well as table-oriented datasets such as WikiTableQuestions \cite{pasupat-liang-2015-compositional}, HybridQA \cite{chen-etal-2020-hybridqa}, FinQA \cite{chen-etal-2021-finqa}, and TAT-QA \cite{zhu-etal-2021-tat}.

More recent benchmarks such as RAGBench \cite{friel-2024-ragbench} and T$^2$-RAGBench \cite{strich-etal-2026-t2} explicitly evaluate retrieval-augmented generation in context-independent settings. However, they still primarily focus on questions grounded within a single document or table instance, even when multiple knowledge types are involved. In contrast, SpecsQA evaluates corpus-level retrieval and reasoning over semi-structured documents, requiring systems to retrieve and combine evidence across multiple documents and content types before answering.

Additional related work appears in Appendix~\ref{appendix:sota}.

\section{SpecsQA Dataset}

SpecsQA evaluates question answering over a corpus of semi-structured product documents, where relevant evidence may appear in natural-language text, specification tables, or across multiple documents. Existing QA benchmarks rarely combine structured and unstructured evidence, and mixed text-table datasets typically focus on questions answerable within a single document. In contrast, SpecsQA targets a corpus-level retrieval setting and it also provides, both natural-language answers and canonical product lists, allowing more reliable evaluation. Moreover, because the benchmark is built from a time-specific snapshot of a commercial website that continuously evolves, it reduces the likelihood that answers can be solved through memorization by pre-trained LLMs alone.

\subsection{Dataset Construction}\label{sec:dataset_generation}

The dataset was created from a snapshot of the Samsung UK online shop (\url{https://www.samsung.com/uk/}) collected on November 14th, 2025. We scraped 2162 webpages spanning 26 categories of products. Product pages combine free-form textual descriptions with structured specification tables, although layouts varied across categories and required dedicated parsing pipelines (more details and examples in Appendix~\ref{appendix:layouts}).

Many products are variants of a broader product family, differing only in features such as color, storage capacity, or screen size. Since multiple variants with different prices are often mapped to the same URL, we additionally stored, besides the raw HTML pages, the variant configuration used when visiting each webpage: we extracted structured metadata including product names, categories, prices, model identifiers, and specification attributes into a unified JSON representation.

\subsection{Questions Design}\label{sec:question_gen}

We manually wrote 117 questions designed to evaluate both retrieval and reasoning over semi-structured product data. Questions were grouped into four categories summarized in Table~\ref{tab:dataset_characteristics}:
\textit{Inverse queries} require retrieving all products satisfying a given property or relation;
\textit{Multi-condition queries} combine several structured constraints simultaneously;
\textit{Group comparison queries} require comparing properties across multiple product families; while
\textit{Reasoning queries} involve more open-ended recommendations and user-preferences.

\begin{table}[h!]
  \caption{Distribution of SpecsQA questions by category, and percentage of objective and list-based answers.}
  \label{tab:dataset_characteristics}
  \resizebox{\linewidth}{!}{%
    \begin{tabular}{lcccc}
      \toprule
      \textbf{Question Category} & \textbf{Proportion} & \textbf{Count} & \textbf{Obj.} & \textbf{List} \\
      \midrule
      Inverse           & $35.0\%$ & $41$ & $100\%$  & $100\%$  \\
      Multi-condition   & $22.2\%$ & $26$ & $100\%$  & $100\%$  \\
      Group Comparison  & $20.5\%$ & $24$ & $100\%$  & \phantom{00}$0\%$  \\
      Reasoning         & $22.2\%$ & $26$ & $7.7\%$ & $96.2\%$ \\
      \midrule
      \textbf{Total}            &  & $117$ & $79.5\%$ & $78.6\%$ \\      
      \bottomrule
    \end{tabular}%
  }
\end{table}

Questions were designed around realistic consumer-oriented criteria such as price ranges, battery capacity, connectivity features, display technology, or device categories. Constraints were intentionally formulated to require either exact filtering or aggregation over structured attributes. For example, a query may ask for all smartphones under a certain price threshold that simultaneously support 5G connectivity, AMOLED displays, and a minimum battery capacity. Additional examples are provided in Appendix~\ref{sub:questionTypes}.

\subsection{Answers Annotation}\label{sec:question_ann}
Ground-truth answers were manually annotated using the information available on the website at the time of scraping. Since product availability and prices evolve over time, we release the original scraped data snapshot together with the benchmark to ensure reproducibility.

Depending on the question type, answers were represented either as natural-language text, product lists, or both. 
Product lists enable deterministic evaluation for questions requiring exact filtering or exhaustive retrieval, while natural-language answers support evaluation of more open-ended reasoning tasks. 
To allow for NL-based evaluation we also automatically generated a NL version of product lists answers using an LLM.
We additionally annotate whether a question is objective or subjective, distinguishing factual specification queries (e.g., which products meet given specifications) from recommendation-oriented questions (e.g., a phone for an elderly person) where multiple answers may be acceptable.

\section{DualGraphRAG}

\begin{figure*}[h]
  \centering
  \includegraphics[width=\linewidth]{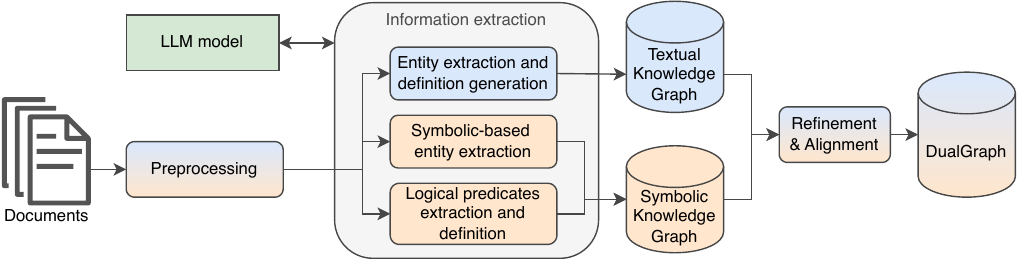}
    \caption{Overview of Dualgraph indexing process. Legend: Blue - TKG processing; Orange - SKG processing.}
    \label{fig:system}
\end{figure*}

We consider question answering over semi-structured corpora containing both natural-language descriptions and structured content such as specification tables. Given a corpus of documents $\mathcal{C}=\{d_i\}_{i=1}^{N}$ and a question $q \in \mathcal{Q}$, the goal is to generate an answer $a$ grounded in the corpus. 
For questions with a well-defined symbolic target (i.e., a canonical set of entities), the answer may also be converted into a symbolic form $a_{\text{sym}}$ (e.g., using an LLM to  extract entities from $a$). We therefore model the task as $ f:\ (q,\mathcal{C}) \mapsto a $
where the system must retrieve a supporting evidence set $e(q,\mathcal{C}) \subseteq \mathcal{C}$ sufficient to generate a complete and grounded answer.

To solve this class of problems, we propose \textit{DualGraph} (see Figure~\ref{fig:system} for an overview) a retrieval framework that converts a corpus $\mathcal{C}$ into an intermediate representation composed of two complementary graph views: a \textit{Textual Knowledge Graph (TKG)} and a \textit{Symbolic Knowledge Graph (SKG)}. The $TKG=(V_T,E_T)$ encodes natural language information as text-annotated nodes and relations, while the $SKG=(V_S,E_S)$ encodes structured information as typed entities and directed predicate edges forming \textit{subject-predicate-object} triples.

\subsection{Textual and Symbolic Graph Views}
The \textbf{Textual Knowledge Graph} is an undirected graph $TKG=(V_T,E_T)$ that captures natural language information in the corpus. Each node $v\in V_T$ is associated with a canonical name and a textual description generated from all the passages supporting the entity, while each edge $e\in E_T$ represents mutual semantic relationships between two entities and is also annotated with a textual description. However, in our implementation we dropped explicit TKG edges and retain only entity nodes for efficiency.
In contrast to the SKG, the TKG preserves linguistic context and ambiguity, making it more suitable for semantic matching, paraphrase robustness, and underspecified questions.

The \textbf{Symbolic Knowledge Graph} is a directed graph $SKG=(V_S,E_S)$ that represents structured facts extracted from the corpus in the form of typed subject--predicate--object triples. 
We assume an ontology-based representation,
where a node $v\in V_S$ corresponds to an individual, a class or a numeric attribute, while each edge $e\in E_S$ represent a logic predicate $p\in\mathcal{P}$ mapping a subject node $v^s$ to an object node $v^o$. 
As any symbolic knowledge graph, SKG supports exact operations including filtering, aggregation, comparison, arithmetic constraints, and exhaustive listing. Questions admitting a symbolic interpretation can therefore be mapped to formal queries (e.g., SPARQL) that, executed directly on the graph, return a set of bindings, triples, or subgraphs, which can be used for answer generation.

\subsection{Information Extraction and Graph Construction}

Given a corpus $\mathcal{C}=\{d_i\}_{i=1}^{N}$ of documents in semi-structured formats, we first decompose them into a sequence of text chunks $t_j$ and a set of semi-structured components $s_j$ (e.g., specification tables), where the chunking granularity is treated as a hyper-parameter. The text chunks are processed into MD files to construct the TKG, while the structured components are processed into JSON files and used to populate the SKG (see Appendix~\ref{appendix:dataset_preprocessing} for more details).

\textbf{Textual Knowledge Graph. } The TKG is induced through open information extraction and does not assume a predefined schema. We follow the method proposed in UnWeaver \cite{tuora2026unweavingknotsgraphrag} which uses an LLM to map each chunk $t_j$ to a set of entity mentions. Entity descriptions are then aggregated across supporting chunks and embedded to form an entity-centric retrieval index (see Appendix~\ref{sec:unweaver} for more details).

\textbf{Symbolic Knowledge Graph.} SKG construction assumes an input ontology $\mathcal{O}$ (i.e., a description-logic schema specifying classes and predicates). In our implementation, we manually designed $\mathcal{O}$ as a lightweight  schema (see Appendix~\ref{appendix:schema}) intended to minimize domain-specific engineering while remaining simple enough to adapt when the underlying data format changes.

The SKG is populated by converting each semi-structured component $s_j$ into ontology-consistent  triples $(v^s,p,v^o)$, where subject $v^s$ and object $s^o$ nodes represent typed entities, attributes, or values, and edges represent predicates from the ontology $\mathcal{O}$. 
Each specification row extracted from a product table is modeled as a \texttt{Spec} entity connected to its product, specification section, feature name, and value node, following the mapping in Appendix~Tables \ref{tab:schema_classes} and \ref{tab:schema_predicates}, Figure \ref{fig:schema}. This preserves the original tabular organization while remaining generic enough to accommodate diverse webpage layouts and product categories. 
Product variants are represented as distinct but related entities connected through shared product-range and category nodes.

Numeric values undergo additional symbolic processing. When possible, numeric quantities are detected within the raw text and converted into typed symbolic representations with normalized units, including the special case of multidimensional quantities where several numerical components must be represented jointly (e.g., ``resolution'' which contains both width and height). This enables exact numerical operations (e.g, filtering with arithmetic constraints) directly in SPARQL. This processing is necessary due to noise in the scraped data, and for the same reason we preserve the original textual value to prevent information loss.

Beyond specification rows, the SKG also models product hierarchies and shared features. 
Products are grouped into ranges and categories, while common features are represented as reusable entities linked to multiple products, reducing duplication and producing more consistent graph patterns for SPARQL generation.

Entity and value normalization is central to SKG construction: product identifiers, feature names, section labels, and categorical values are canonicalized through lightweight heuristics including lowercasing, stemming, stop-word removal, and string normalization, reducing duplication due to inconsistent webpages or table structures.

The SKG is then enriched via lightweight rule-based reasoning. We defined a set of Datalog rules (see Appendix~\ref{appendix:datalog}) on top of the schema $\mathcal{O}$ to derive higher-level features implicitly expressed in the specifications (e.g., inferring ``5G Support'' from the presence of specific 5G standards).

Finally, the two graph views are refined through alignment: We first apply joint canonicalization heuristics that normalize identifiers, names, and types, and merge duplicate nodes. This simple step already aligns many entities across $V_T$ and $V_S$, since many identifiers in e-commerce corpora are already partially standardized. We also experimented with a learned contrastive alignment model based on textual and graph embeddings, but observed no measurable improvement.

\subsection{Retrieval and Question Answering}

Given a question $q$, DualGraph supports two main retrieval functions: a symbolic SKG-based retriever, implemented through a LLM-based SPARQL query generation, and a semantic TKG-based retriever, following the entity-centric retrieval procedure of UnWeaver \cite{tuora2026unweavingknotsgraphrag}.

\textbf{Semantic Retrieval.} 
The retrieval function used in UnWeaver is similar to VectorRAG, but uses entities as an intermediate layer between questions and chunks. Given a question $q$, the system first retrieves the entities whose descriptions are most similar to $q$. Each retrieved entity then votes for its originating chunks, with votes weighted by the entity's similarity rank. This produces a final chunk ranking, from which the top-$k$ are selected.

\textbf{Symbolic retrieval.~}
Symbolic retrieval maps a natural-language question $q$ to a SPARQL query executed over the SKG. The NL-to-SPARQL translation is performed by an LLM prompted with $q$ and two context components: a \textit{common component} shared across all questions, and a \textit{question-specific component} retrieved at inference time.

The \textit{common component} specifies: (1) a persona defining the model as an expert in SPARQL, RDF, OWL, and related standards; (2) query-generation rules for selecting query types, applying filters, and incorporating constraints; (3) a domain description of online-shop product data represented as a knowledge graph; (4) the SKG schema, including rules for handling numerical values and prices; and (5) the expected output format.

The \textit{question-specific component} consists of retrieved graph patterns. After graph construction, predefined schema-based patterns are linearized automatically into natural language, embedded, and stored in a vector database. At inference time, the patterns most similar to $q$ are retrieved and used to ground SPARQL generation. We use four pattern types: \textit{Spec patterns}, describing product attributes (e.g., \textit{In the product specification, the S Pen Support entry in the Specifications section has the value Yes}); \textit{Feature patterns}, describing product capabilities (e.g., \textit{The product has Samsung Dex Support feature}); \textit{Category patterns}, representing product-family information (e.g., \textit{Galaxy S}); and \textit{Singular Node patterns}, representing individual entities appearing in triples.

In our experiments, we retrieve $5$ instances of each pattern type and convert them into syntactically valid SPARQL snippets, which are included in the LLM prompt. The LLM then generates candidate SPARQL queries, which are executed over the SKG. Results are formatted as Markdown tables together with the query that produced them. We discard queries that fail execution, return more than $100$ results, or return no results when all other candidates also fail or are discarded. 

To improve robustness, we generate multiple candidate SPARQL queries ($3$ in our experiments) through controlled sampling, execute them on the SKG, and use all the resulting outputs as context for answer generation.

We also experimented with an agentic symbolic retrieval, where an LLM iteratively refines SPARQL queries based on execution feedback. However, since it did not improve retrieval performance, we do not report these results.

\subsection{Orchestration and Answer Generation}

DualGraph supports multiple orchestration strategies combining symbolic and semantic retrieval (see Appendix-Figure~\ref{fig:query_diagram}). We evaluate seven variants: \textbf{(1)} \textit{TKG only}, which applies semantic retrieval only; \textbf{(2)} \textit{SKG only}, which applies symbolic retrieval only; \textbf{(3)} \textit{TKG + SKG}, which concatenates the contexts returned by both retrievers; \textbf{(4)} \textit{SKG + TKG fallback}, which applies semantic retrieval when symbolic retrieval returns no context; \textbf{(5)} \textit{Router}, which uses an LLM binary classifier to select between symbolic and semantic retrieval conditioned on the user question, descriptions of the retrieval methods, and few-shot examples; \textbf{(6)} \textit{Router + TKG fallback}, which applies semantic retrieval as fallback when routed symbolic retrieval fails; and \textbf{(7)} \textit{Agentic}, which uses an LLM agent with access to both retrieval functions.
The agentic orchestrator, implemented with Pydantic-AI \cite{pydantic-ai:online}, iteratively decides whether to invoke a retriever, reformulate the query, generate an answer, or continue retrieval based on intermediate outputs. After generating an answer, it self-reflects and either returns the answer, revises it, or continues the loop until a stopping criterion is reached.

Conditioned on the retrieved context and $q$, a generator LLM produces a natural-language answer $a$. When symbolic output is required for evaluation, an additional LLM-generation step extracts a canonical symbolic answer $a_{\text{sym}}$ from $a$.

\begin{table*}[t]
  \caption{Comparison between DualGraph and state-of-the-art baselines on SpecsQA. We report answer quality using factual correctness (FC), list matching (LM), pairwise LLM-as-a-judge (LaaJ), and objective-only LM, together with indexing and query-time token usage.}
  \label{tab:results_dualgraph_vs_sota}
  \centering
  \resizebox{0.9\linewidth}{!}{%
    \begin{tabular}{lccccS[table-format=3.3]S[table-format=3.3]}
      \toprule
        & \multicolumn{4}{c}{\textbf{Answer Quality}} & \multicolumn{2}{c}{\textbf{Token Usage}} \\
       \cmidrule[0.5pt](lr{0.3em}){2-5} 
       \cmidrule[0.5pt](lr{0.3em}){6-7} 
       & \textbf{FC (F1)} & \textbf{LM (F1)} & \textbf{LaaJ}  & \textbf{Obj. LM (F1)}  & \textbf{Indexing ($\cdot 10^6$)} & \textbf{Query ($\cdot 10^3$)}\\
      \midrule
      LLM only 
            & 0.107  & 0.028 & 0.559
            & 0.036  
            & 0 & 0.565 \\
      Vector RAG 
            & 0.092 & 0.118 & 0.438
            & 0.158  
            & 0 & 2.436 \\
      Microsoft GraphRAG 
            & 0.153  & 0.203  & 0.575
            & 0.243  
            & 181.213  & 11.120 \\
      Microsoft GraphRAG (fast) 
            & 0.083  & 0.140 & 0.420
            & 0.158  
            & 29.420 & 12.236 \\
      RAPTOR 
            & 0.207 & 0.216  & 0.568
            & 0.248 
            & 5.615 & 2.378 \\
      LinearRAG 
            & 0.138  & 0.085  & 0.526
            & 0.093 
            & 0 & 7.541 \\
      AriGraph 
            & 0.072 & 0.124 & 0.378
            & 0.140 
            & 16.516 & 5.097 \\
      HippoRAG 2 
            & 0.149  & 0.152 & 0.540
            & 0.182 
            &16.084 & 22.217 \\
      Wikontic 
            & 0.135  & 0.290 & 0.523
            & 0.349  
            & 420.325  & 138.627 \\
      A-RAG 
            & 0.057  & 0.129  & 0.331
            & 0.159 
            & 0 & 36.384 \\
      TableRAG 
            & 0.091 & 0.118 & 0.453 
            & 0.139  
            & 0 &  20.527 \\
      \bf DualGraph (R+TKG\_fb) (our) 
            & \bf 0.298  & 0.357 & 0.640
            & \bf 0.441  
            &  16.058 & 7.678 \\
      \bf DualGraph (SKG+TKG\_fb) (our) 
            & 0.293  & \bf 0.372 &  \textbf{0.644}
            &  0.434 
            & 16.058 & 6.802 \\
      \bottomrule
    \end{tabular}%
  }
\end{table*}

\begin{table}
  \caption{Ablation of our different DualGraph variants.}
  \label{tab:ablation_dualgraph_variants}
  \resizebox{\linewidth}{!}{%
        \begin{tabular}{lcccS[table-format=2.3]}
      \toprule
        & \multicolumn{3}{c}{\textbf{Answer Quality}} & {\textbf{Token Usage}} \\
       \cmidrule[0.5pt](lr{0.3em}){2-4} 
       \cmidrule[0.5pt](lr{0.3em}){5-5} 
      \bf DualGraph  & \textbf{FC (F1)} & \textbf{LM (F1)} & \textbf{LaaJ}  & \textbf{Query ($\cdot 10^3$)}\\
      \midrule      
      SKG only                         
            & 0.273 & 0.321 & 0.507 & 5.471 \\
      TKG only                                  
            & 0.140 & 0.136 & 0.352 & 3.095 \\
      SKG concat TKG                 
            & \bf 0.306 & 0.367 & 0.561 & 8.186 \\
      \bf  SKG+TKG\_fb               
            & 0.293  & \bf 0.372  & 0.523 & 6.802 \\
      Router                         
            & 0.268  & 0.303 & 0.485 & 6.503 \\
      \bf Router+TKG\_fb             
            & 0.298 & 0.357 & 0.528 & 7.678 \\
      Agentic                 
            & 0.240  & 0.341 & \textbf{0.764} & 55.454 \\

      \bottomrule
    \end{tabular}%
  }
\end{table}

\section{Results}\label{sec:results}

We performed several experiments of which the main results can be summarized as follows:
\textbf{1)} DualGraph outperforms state-of-the-art baselines, with the largest gains on specification-heavy queries requiring exact filtering and exhaustive lists;
\textbf{2)} The SKG--TKG duality is key for handling diverse question types: symbolic retrieval improves precision over structured attributes while textual retrieval improves coverage and robustness;
\textbf{3)} SpecsQA serves as a  diagnostic benchmark for semi-structured QA, revealing the strengths and weaknesses of different retrieval strategies.
The full set of results, together with additional experiments, can be found in Appendix~\ref{appendix:full_results}.

\subsection{Baselines and Metrics}\label{sec:metrics}
We compared our method with state-of-the-art baselines including: pure LLM calls (`\textit{LLM-only}'), standard vector-based RAG systems (`\textit{Vector RAG}') and state-of-the-art graph-based (`\textit{Microsoft GraphRAG}' \cite{edge2025localglobalgraphrag} -- both standard and fast version, `\textit{RAPTOR}' \cite{DBLP:conf/iclr/SarthiATKGM24}, `\textit{LinearRAG}' \cite{zhuang2025linearraglineargraphretrieval}, `\textit{AriGraph}' \cite{ijcai2025p2}, `\textit{HippoRAG~2}' \cite{gutierrez2025hipporag2}), agentic (`\textit{A-RAG}' \cite{a-rag2026}), symbolic (`\textit{Wikontic}' \cite{chepurova2025wikontic} ) and table-oriented (`\textit{TableRAG}' \cite{yu2025tableragretrievalaugmentedgeneration}) RAG methods.

We evaluate all systems with four metrics:
\textit{Factual Correctness} from RAGAS \cite{ragas2024} measures whether generated statements are supported by the ground truth, after decomposing both into atomic claims;
\textit{List Match} evaluates list-answer questions by extracting (with an LLM) a predicted symbolic list from the generated answer and computing set-based precision, recall, and F1 against the ground-truth list (when available);
\textit{Pairwise LLM-as-a-judge (LaaJ)} compares system outputs in pairs using a fixed prompt and reports each system's aggregate win rate; and
\textit{Computational cost}, measured as input and output token usage.

For all experiments, we used \texttt{GPT-OSS-120B} as the underlying LLM, \texttt{Qwen3-Embedding-4B} as embedding model and report results averaged over 5 indexing runs, 3 query-generation runs, and 3 evaluation runs to reduce stochastic variability.

\subsection{DualGraph vs Baselines}
Table~\ref{tab:results_dualgraph_vs_sota} compares DualGraph with the considered RAG baselines. We report our two most representative variants:
\textit{Router+TKG fallback} achieves the highest factual correctness, and \textit{SKG+TKG fallback} achieves the best list matching and LLM-as-a-judge scores. The strongest non-DualGraph baselines are RAPTOR for factual correctness and Wikontic for list matching.

Overall, the results show that combining symbolic querying over structured specifications with semantic retrieval substantially improves QA quality on semi-structured corpora. Text-only methods retrieve locally relevant passages but struggle with global questions, while graph-based baselines improve retrieval structure but still lack support for exhaustive lists and precise filtering. Wikontic remains competitive on symbolic questions, but at significantly higher indexing and query cost.

DualGraph provides a favorable quality--cost trade-off: its indexing cost is comparable to HippoRAG~2 and AriGraph, and substantially lower than Microsoft GraphRAG and Wikontic, while maintaining moderate query-time cost. 
Finally, the relatively low list-matching scores across systems indicate the difficulty of SpecsQA, where answers are sensitive to indexing noise and product availability. Moreover, we observe consistently higher performance on objective questions, since subjective recommendation-oriented queries may admit multiple valid answers that may differ from the ground-truth annotations.

\begin{figure*}[!t]
  \centering
  \includegraphics[width=\linewidth]{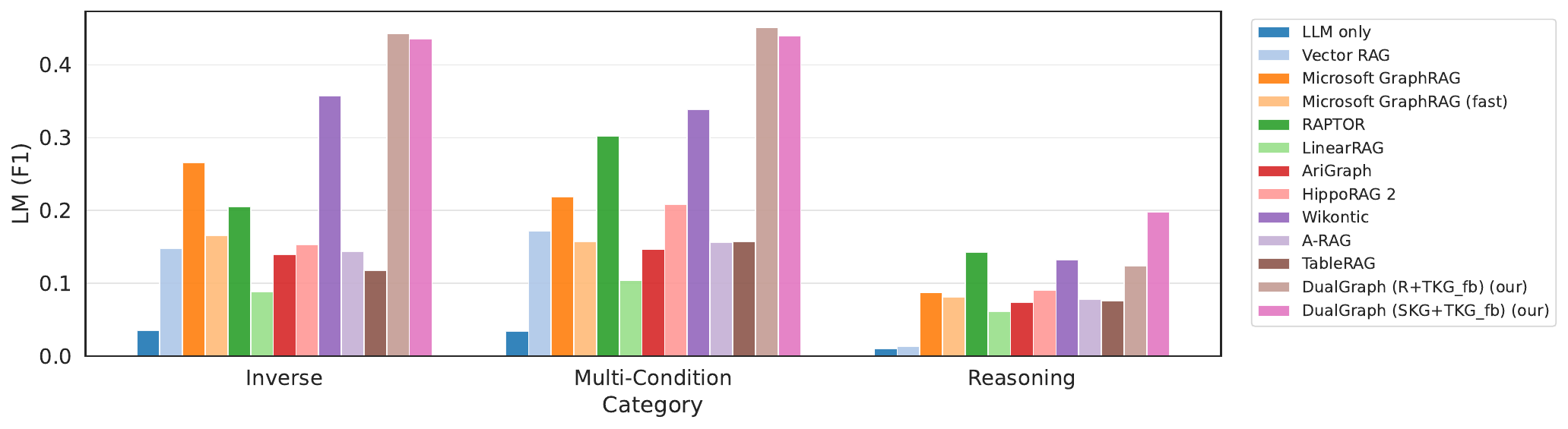}
  \caption{List matching (F1) results by question category for DualGraph and baseline retrieval methods.}
  \label{fig:categories_list_matching_f1}
\end{figure*}

\subsection{Impact of SKG and retrieval strategies}\label{sec:ablation}
Table~\ref{tab:ablation_dualgraph_variants} reports an ablation over DualGraph retrieval strategies. 
TKG-only retrieval performs substantially worse than SKG-based variants, especially on list matching, confirming that semantic retrieval alone is insufficient for semi-structured QA. Hybrid variants consistently outperform single-view retrieval, showing that symbolic and textual representations provide complementary strengths.

Among the variants, \textit{SKG+TKG fallback} achieves the best list-matching performance, while \textit{Router+TKG fallback} offers the best general-purpose trade-off between performance and token usage by selecting retrieval strategies dynamically and falling back to textual retrieval when symbolic querying fails. The former is particularly effective on SpecsQA because many questions are specification-heavy, whereas the latter is likely preferable for more balanced datasets. Concatenating both contexts slightly improves factual correctness but increases query cost, while the agentic router is substantially more expensive due to iterative retrieval steps. However, it also achieves the highest LLM-as-a-judge score, possibly reflecting known verbosity and superficial-quality biases in LLM-based evaluators \cite{zhou2024mitigating}.

Overall, the results support the central design of DualGraph: symbolic retrieval improves precision on structured constraints, while textual retrieval increases robustness on open-ended and underspecified questions.

\subsection{Impact of Question Categories}

Figure~\ref{fig:categories_list_matching_f1} reports performance by question category. DualGraph performs
particularly well on inverse and multi-condition queries, where answers usually require exact filtering and exhaustive product retrieval. These gains mainly derive from the SKG component and its support for structured querying. These results demonstrate the value of the SKG component for specification-heavy questions.

Performance is lower across all systems on group-comparison (see Appendix Fig \ref{fig:category_full} and \ref{fig:category_ablation_full}) and reasoning questions, especially for subjective recommendation-style queries. In these cases, the TKG provides complementary natural-language context that improves robustness beyond purely symbolic retrieval.
The category-level breakdown also shows the diagnostic value of SpecsQA, making it possible to identify which retrieval capabilities are required by different classes of questions.

\subsection{Impact of Patterns}

Table~\ref{tab:ablation_patterns} reports an ablation on the contribution of graph patterns to SPARQL generation quality. 
\textit{Spec}  appears to be the most critical pattern: removing it leads to the largest drop in successful SPARQL retrievals and substantially lowers both factual correctness and list matching. This shows that detailed specification-level descriptions are essential for constructing accurate symbolic queries. The fact that using the \textit{Spec} pattern alone remains competitive further confirms its foundational role in symbolic retrieval.
\textit{Feature} and \textit{Singular Node} patterns provide smaller but consistent improvements, suggesting that they offer complementary context for some question types. The \textit{Category} pattern appears less beneficial and may even degrade performance: the best overall performance is achieved when it is removed while all other patterns are retained. 
However, we keep all patterns in the main experiments because the full configuration achieves the highest SPARQL retrieval success rate and robust performance across metrics. The marginal gains from removing Category are outweighed by the benefit of a more general default that can adapt to diverse question types beyond SpecsQA.

\begin{table}[t]
  \caption{Ablation of graph patterns [Spec, Feature, Category, Singular Node] used for SPARQL generation in DualGraph (SKG with TKG fallback). We report the percentage of successful SKG SPARQL retrievals, factual correctness (FC), list matching (LM), and LLM-as-a-judge (LaaJ) scores. }
  \label{tab:ablation_patterns}
  \centering
  \resizebox{0.9\linewidth}{!}{%
    \begin{tabular}{ccccc}
      \toprule
    \textbf{Pattern} & \textbf{\% SKG} & \textbf{FC(F1)} & \textbf{LM(F1)} & \textbf{LaaJ} \\
      \midrule
      \blue{Y  Y  Y  Y}       &  \bf 50.2  & 0.293 & 0.372 & 0.534 \\
      N \blue{ Y  Y  Y}       &  37.7  & 0.247 & 0.292 & 0.461 \\
      \blue{Y} N  \blue{Y  Y} &  48.4  & 0.299 & 0.361 & 0.521 \\
      \blue{Y  Y}  N \blue{Y} &  48.8  & \bf 0.317 & \bf 0.380 & 0.538 \\
      \blue{Y  Y  Y}  N       &  48.8  & 0.296 & 0.367 & 0.536 \\
      \blue{Y}  N  N  N       &  44.7  & 0.301 & 0.362 & \bf 0.545 \\
      N  \blue{Y}  N  N       &  31.3  & 0.246 & 0.290 & 0.478 \\
      N  N  \blue{Y}  N       &  34.7  & 0.228 & 0.259 & 0.449 \\
      N  N  N  \blue{Y}       &  37.0  & 0.246 & 0.299 & 0.472 \\
      N  N  N  N              &  29.4  & 0.230 & 0.253 & 0.459 \\
      \bottomrule
    \end{tabular}%
  }
\end{table}

\subsection{Conclusions}

To conclude, we introduced DualGraph, a RAG architecture that leverages the strengths of both semantic and symbolic retrieval, combining flexible language-based retrieval with exact structured querying, and outperforming state-of-the-art baselines particularly on questions requiring precise filtering and exhaustive retrieval.
We also presented SpecsQA, a useful new benchmark for question answering over semi-structured documents, a common scenario in real-world applications.

\clearpage

\section*{Limitations}

Our work opens several directions for further improvement. 
In the current implementation, the alignment between the TKG and SKG is lightweight: because symbolic nodes do not have textual descriptions, SKG and TKG entities are mainly aligned through normalized names. However, even this simple alignment allows the two graph views to provide complementary retrieval functions. Richer cross-graph alignment could further exploit this complementarity, enabling genuinely joint SKG--TKG retrieval in which symbolic results are expanded with textual evidence, or semantic retrieval is constrained by symbolic structure. Improving this alignment through richer node descriptions, an improved learned cross-graph matching, or retrieval functions that explicitly traverse both views is therefore a promising direction for future work.

The SKG is currently constructed mainly from the structured portions of the corpus, which in SpecsQA correspond to product specification tables. This choice provides a clean and reliable symbolic representation for attributes that are naturally tabular, and is well suited to exact filtering, comparison, and list retrieval. Some product information, however, may appear only in natural-language descriptions, marketing text, or FAQs. Extending symbolic extraction to these textual sections, or using stronger SKG--TKG alignment to combine partial symbolic evidence with textual context, would allow DualGraph to cover an even broader range of semi-structured questions.

Several symbolic components, including the initial ontology, SPARQL retrieval patterns, and Datalog rules, are currently specified by a domain expert. This manual design makes the pipeline interpretable, controllable, and robust in our setting, while keeping the symbolic representation compact. Future work could reduce this domain-specific effort by automatically inducing schemas and rules, learning derived features from data, or extracting useful SPARQL patterns from the graph schema and query distribution.

Finally, our orchestration strategies are deliberately simple and general-purpose. We use LLM-based routing and agentic retrieval variants to study how symbolic and textual retrieval can be combined without task-specific training. More adaptive routing policies could further improve performance, especially for ambiguous questions or cases requiring close interaction between both graph views. For example, a router trained on task-specific feedback could learn when to use symbolic retrieval, textual retrieval, or a joint strategy.

\clearpage
\onecolumn

\appendix

\section{Extended Literature Review}\label{appendix:sota}

\subsection{Graph RAG}
GraphRAG methods extend standard retrieval-augmented generation by converting a document corpus into a knowledge graph structure, where entities are represented as nodes and relations as edges, with links to supporting passages. At inference time, relevant subgraphs are selected using semantic similarity or graph traversal, enabling the language model to synthesize answers from connected evidence across multiple documents.

Microsoft GraphRAG \cite{edge2025localglobalgraphrag} constructs a knowledge graph from source documents by extracting entities, relationships, and claims using large language models. The method partitions the graph into hierarchical communities and generates community-level summaries that enable both global sensemaking queries and local retrieval operations. GraphRAG offers two indexing variants: \textit{standard}, which uses LLM-based extraction for rich entity and relationship descriptions, and \textit{fast}, which employs traditional NLP techniques for faster and cheaper indexing. While GraphRAG effectively supports multi-document queries through hierarchical summarization, its graph representation is primarily textual and does not support exact symbolic operations or formal querying over structured attributes.

RAPTOR \cite{DBLP:conf/iclr/SarthiATKGM24} builds a hierarchical index through recursive embedding, clustering, and summarization of text chunks using Gaussian Mixture Models. The method constructs a multi-layered tree structure from the bottom up, where each level represents increasingly abstract summaries of the underlying content, enabling retrieval at varying granularities. RAPTOR's collapsed tree querying method allows simultaneous evaluation across abstraction levels. However, RAPTOR operates on hierarchical text summaries rather than a structured graph, and thus cannot support precise filtering or aggregation over typed attributes.

LinearRAG \cite{zhuang2025linearraglineargraphretrieval} addresses the instability of traditional relation extraction by constructing a relation-free hierarchical graph comprising entity, sentence, and passage nodes connected through contain and mention adjacency matrices. The framework employs a two-stage retrieval mechanism with local semantic bridging followed by global importance aggregation using personalized PageRank. While LinearRAG achieves efficient linear-time complexity, its relation-free representation prevents exact symbolic querying and structured reasoning over predicates.

AriGraph \cite{ijcai2025p2} integrates semantic and episodic memories within a unified graph structure for LLM agents. The system continuously extracts semantic triplets from textual observations and links them with episodic vertices storing raw observations. While AriGraph demonstrates competitive performance on multi-hop question-answering benchmarks at lower computational cost than dedicated knowledge graph methods, it remains focused on semantic retrieval rather than supporting formal symbolic operations over structured data.

RAGonite \cite{roy2024ragoniteiterativeretrievalinduced} is a retrieval-augmented generation system designed for conversational question answering over RDF knowledge graphs. The method employs a two-pronged retrieval approach that fuses SQL query results over a database automatically derived from the KG with text-search results over verbalizations of KG facts. RAGonite supports iterative retrieval, enabling multiple rounds of evidence gathering when initial results are unsatisfactory, and integrates both retrieval branches through an LLM to generate coherent responses. This design shares conceptual similarities with DualGraph in maintaining complementary symbolic and textual representations. However, RAGonite assumes a pre-existing structured knowledge graph as input and operates over a database representation, whereas DualGraph constructs both aligned graph views from unstructured and semi-structured documents. This enables DualGraph's entity-centric retrieval to leverage graph connectivity and expand evidence through neighborhoods over the aligned graphs, while RAGonite's SQL-based approach is limited to relational queries without access to the underlying graph structure.

HippoRAG 2 \cite{gutierrez2024hipporag,gutierrez2025hipporag2} is an advanced retrieval-augmented generation framework inspired by the neural mechanisms of human long-term memory. An LLM is presented as an artificial neocortex, a knowledge graph (KG) with Personalized PageRank (PPR) algorithms is shown as analogous to the hippocampus, and a retrieval encoder maps to the temporal lobe. Unlike traditional RAG methods that rely solely on vector-based retrieval, HippoRAG 2 enhances the search process by incorporating KG triples, enabling deeper contextualization and memory-aware filtering. During offline indexing, the LLM converts documents into KG triples, while the retrieval encoder detects synonyms. At inference time, the system extracts entities from queries, performs context-aware searches using PPR, and filters results through a two-stage recall-and-recognition process. HippoRAG 2 introduces dense-sparse integration to minimize information loss, and deeper contextualization to improve search accuracy. While the framework excels in factual memory, sensemaking, and associative tasks (outperforming RAG baselines by 7\% in associative tasks) it retains the limitations of graph-based approaches, such as the inability to support exact symbolic operations over structured attributes. Its computational efficiency and robust performance across diverse benchmarks position it as a significant advancement in non-parametric continual learning for LLMs. The concept of embedding whole triples is similar to our embedding of patterns, but the application is completely different, as in our system the patterns provide anchors for querying the SKG, while in HippoRAG they drive textual retrieval. Our patterns are also more complex and linearized in a more natural way.

\subsection{Symbolic RAG}
Symbolic RAG is a subset of GraphRAG methods that encode the corpus with formal semantics, typically using description logics, ontologies, and typed predicates, by mapping documents into a logical knowledge graph where nodes denote typed entities and edges denote predicates between them. 
In many systems~\cite{chepurova2025wikontic,mo2025kggen}, retrieval follows the standard GraphRAG pipeline, selecting relevant subgraphs via semantic similarity and optionally expanding them through graph traversal to collect more evidence. 
In other approaches~\cite{smeros2025sparqlllmrealtimesparqlquery,emonet2025llm,ragonite}, inference relies on translating the question into a formal query, such as SPARQL or an equivalent logic form, to retrieve a set of triples or a subgraph, which is then returned directly or provided as context to a language model for natural language generation.

Symbolic RAG includes a subclass of RAG systems suited especially for answering questions on data in tabular format. Recent RAG pipelines target multi-table QA by combining retrieval of relevant tables with structured reasoning. For example T-RAG \cite{zou2025rag}, a table-corpora-aware RAG framework builds a hierarchical index over a large collection of tables and uses multi-stage retrieval with a graph-aware prompt to aggregate evidence across tables. Similarly, OpenTab \cite{kong2024opentab} retrieves relevant tables and uses LLM-generated SQL programs to parse these tables, yielding more accurate answers.
TableRAG \cite{yu-etal-2025-tablerag} loads tables into a database and iteratively performs query decomposition, text retrieval, SQL execution, and answer generation. It achieves state-of-the-art results on the new HeteQA benchmark, confirming the value of SQL-based, global reasoning over tables.
All these methods explicitly preserve table semantics, addressing issues with simply flattening tables into text that often “compromises” structure and hinders multi-hop reasoning \cite{yu2025tablerag}.

\subsection{Agentic RAG}
Agentic RAG systems augment standard retrieval-plus-generation pipelines with planning, modular reasoning, tool usage and self-reflection \cite{li2025survey,singh2025agentic}. 
For instance, CogPlanner \cite{cogplanner2025} adds an iterative planning module for multimodal RAG, significantly improving accuracy and efficiency.
A-RAG \cite{a-rag2026} features hierarchical retrieval interfaces that enable LLMs to autonomously access corpus information at keyword, sentence, and chunk levels enabling the agent to adaptively search and retrieve information at different levels of granularity.
ARAG \cite{arag2025} employs specialized agents (user understanding, NLI inference, context summarization, and item ranking) to refine retrieval for personalized recommendation, achieving large improvements.
MAIN-RAG \cite{chang2025main} introduces a training-free multi-agent pipeline that dynamically filters and scores retrieved documents via agent consensus.
RAGentA \cite{ragenta2025} similarly uses a multi-agent workflow with hybrid sparse+dense retrieval and inline citations to enhance answer faithfulness.
MA-RAG \cite{nguyen2025marag} decomposes multi-hop QA via distinct planner, extractor, and answer-generation agents (with chain-of-thought prompting), enabling even small LLMs to match larger models and setting new SOTA on complex QA.
CoT-RAG \cite{li2025cot} integrates knowledge-graph-driven CoT and pseudo-program prompts within a RAG pipeline, yielding substantial accuracy gains across diverse reasoning benchmarks.
For multimodal QA, HM-RAG \cite{liu2025hmrag} uses a hierarchical agent stack (decomposition, multi-source retrieval, decision) over text, graph, and web data. 
Agent-as-a-Graph \cite{nizar2025agentag} frames agent/tool selection as a knowledge-graph retrieval problem, improving relevant-agent recall in multi-agent RAG settings.

\subsection{RAG Datasets.~}

RAG QA systems have been evaluated on a range of datasets spanning pure text, knowledge graphs, and tables. 
Popular text-only QA benchmarks include open-domain tasks like TriviaQA \cite{joshi-etal-2017-triviaqa}, as well as multi-hop textual QA such as HotpotQA \cite{yang-etal-2018-hotpotqa} and MuSiQue \cite{trivedi-etal-2022-musique}. All of the above contain questions which require reasoning across multiple paragraphs, with answers found in multiple documents (global retrieval). Note that these datasets share a common weakness -- are based on Wikipedia, which means that LLMs were pre-trained on their content.

Separate group of datasets evaluate answering questions about data contained in knowledge graphs. Spider \cite{yu-etal-2018-spider} is a cross-domain text-to-SQL data set containing 10,181 questions on 200 databases. Questions require generating SQL against the provided database schema (local context).
WebQuestions \cite{berant-etal-2013-semantic} is a QA dataset with 6 thousand entries using the Freebase knowledge graph. Questions are answered by querying the entire KG (global context).
ComplexWebQuestions \cite{talmor-berant-2018-web} is another QA corpus with $\sim$34K questions combining Web search and Freebase. Each question is decomposed into subquestions using the full web/KG (global retrieval). 
The 2WikiMultiHopQA corpus \cite{2WikiMultiHopQA} integrates structured knowledge by using Wikidata: questions require combining information from a Wikidata subgraph and associated Wikipedia text.

Another group of datasets focuses on tabular data. Many of them are also based on Wikipedia contents.
KET-QA \cite{hu-etal-2024-ketqa} is a knowledge-enhanced table QA dataset containing 9.4 thousand questions that link each Wikipedia table to a subgraph of Wikidata. Questions require combining table values with graph facts, with local context per question.
Similarly, HybridQA \cite{chen-etal-2020-hybridqa} is a larger, hybrid table-text QA dataset (70K examples) where each question is paired with a Wikipedia table and linked passages. Answers require combining table and text, so each query uses the provided local context. 
WikiTableQuestions \cite{pasupat-liang-2015-compositional} takes a simpler approach -- it is a table QA dataset (22K) where each question is answered using a single HTML table. It contains no extra text; each query is local to its table.
WikiSQL \cite{zhong2017seq2sqlgeneratingstructuredqueries} is another text-to-SQL dataset, containing 80,654 questions over 24,241 Wikipedia tables. Questions are answered by generating SQL on the local table, no extra text is provided.

While Wikipedia is by far the dominant source of knowledge for such datasets, it is not the only one. Quite naturally, the tabular structure of data is most common in financial contexts, resulting in another class of datasets.
For example, FinQA \cite{chen-etal-2021-finqa} is a financial QA dataset with 8.3K questions on financial reports. Each question comes with a PDF-derived table and accompanying text. Answers require numeric reasoning over this local context, making this dataset more specialized.
TAT-QA \cite{zhu-etal-2021-tat} is a financial hybrid QA dataset of 16.5K questions on annual reports. Questions involve both table and text content from the report and each question uses its associated table and/or text (local context).

In such benchmarks, questions generally require integrating information drawn from a particular table or graph instance.
These datasets are not well suited for evaluating retrieval-augmented generation (RAG) systems because the questions are inherently context-dependent: they are unambiguously answerable only given the pre-specified table or graph. Their primary purpose is to assess reasoning and answer generation conditioned on a provided context, rather than to assess the quality or effectiveness of retrieval mechanisms.

The gap is not absolute though -- recently, some dedicated RAG benchmarks have been introduced. T$^2$-RAGBench \cite{strich-etal-2026-t2} is a hybrid financial benchmark (text+tables) of 23,088 QA pairs designed for RAG evaluation on real-world documents and created by concatenating FinQA, TAT-QA and ConvFinQA \cite{chen-etal-2022-convfinqa}. While in the source datasets each question is tied to its specific table and text (local context), here that oracle context isn't explicitly provided; all examples have been transformed to require retrieval of that context. In fact, 91.3\% of the dataset consists of expert-verified context-independent questions. However, the dataset does not include questions that draw on multiple sources or demand combining or summarizing information across several documents.
Outside of the financial niche, RAGBench \cite{friel-2024-ragbench} offers $\sim$100K examples of industrial text QA (user manuals), which explicitly evaluate end-to-end retrieval+generation. It spans multiple domains and assumes a global corpus (no per-question context).

\section{Additional Implementation Details}

For all experiments, both for our and the baselines methods, we used \texttt{GPT-OSS-120B} \cite{openai2025gptoss120bgptoss20bmodel} as the underlying LLM, \texttt{Qwen3-Embedding-4B} \cite{qwen3embedding} as embedding model and report results averaged over 5 indexing runs, 3 query-generation runs, and 3 evaluation runs to reduce stochastic variability.

\subsection{Baselines methods}

We maintained the default hyperparameter configurations for each baseline method unless otherwise specified, with modifications documented only when necessary to ensure stable execution or to align with official implementation guidelines.

\textbf{Microsoft GraphRAG.} \cite{edge2025localglobalgraphrag} For the fast variant of GraphRAG, we followed the official documentation recommendations and configured the chunk size to 100 tokens with a 15-token overlap\footnote{\url{https://microsoft.github.io/graphrag/index/methods/\#fastgraphrag}}. This configuration ensures optimal performance while maintaining the system's graph-based retrieval capabilities.

\textbf{RAPTOR.~} \cite{DBLP:conf/iclr/SarthiATKGM24} We encountered numerical instability during the Gaussian Mixture Model clustering phase, which resulted in ill-defined empirical covariance due to singleton or collapsed samples. Despite converting input data to float64 precision as recommended, the issue persisted. To address this, we modified several hyperparameters: we increased the chunk size from $100$ to $200$ tokens, raised the generated summary token limit from $100$ to $150$ tokens, and expanded the total cluster length limit from $3500$ to $10000$ tokens. These adjustments provided more robust data for clustering by ensuring larger, more informative text segments. Additionally, we modified the summarization prompt to \textit{"Write a short (up to \{max\_tokens\} words, can be less, do not count words) summary of the following, including as many key details as possible: \{context\}."} This change was necessary because the original token-based limit conflicted with the thinking tokens generated by the language model, causing unpredictable output lengths. These modifications maintained RAPTOR's core hierarchical clustering approach while ensuring stable execution in our experimental setup.

\textbf{Wikontic.~} \cite{chepurova2025wikontic}  
Wikontic processes the datasets at the level of individual files, while we study RAG in a ``corpus setting'', where evidence can be distributed across documents and retrieval is performed over the whole corpus at the same time. This distinction matters because file-level question answering implicitly provides part of the solution, namely which file contains the answer, and is therefore a much simpler task.

Running WikOntic on our full corpus as a single file is not practical, since it splits text into sentences/chunks and processes them sequentially, which would be time-wise prohibitive at our scale. Instead, we extract graphs independently for each document (after chunking, as in the other baselines) and then merge them into a single graph. Merging uses lightweight heuristics for entity matching, such as stemming and stop-word removal. We do not apply additional canonicalization beyond these heuristics.

Also, the retrieved triples generate a context that goes beyond the limit for gpt-oss-120B of 131k tokens. Therefore we set a cap on this context at 130k.

\textbf{LinearRAG} \cite{zhuang2025linearraglineargraphretrieval}, 
\textbf{AriGraph} \cite{ijcai2025p2},
\textbf{HippoRAG 2} \cite{gutierrez2024hipporag,gutierrez2025hipporag2},
\textbf{A-RAG} \cite{a-rag2026}, and
\textbf{TableRAG} \cite{yu-etal-2025-tablerag} required no modifications.

\subsection{Dataset Preprocessing}\label{appendix:dataset_preprocessing}

The raw dataset we released consists of a set of scraped HTML files and metadata in JSON format. While this raw representation can support different downstream uses, in our experiments we first applied two  preprocessing stages to convert it into the textual $t_j$ and structured $s_j$ inputs required by our pipeline.

The first phase (\verb|parse.py|) extracts structured data from HTML. The parsed tables, corresponding to 2327 individual products in total, are stored in a single \verb|specs.json| file as a list of dictionaries of $attribute:value$ pairs  -- note that attributes were often structured, e.g. `$Display.Resolution (Main Display): 2160\times1856 (QXGA+)$'.

The second pre-processing phase (\verb|dataset.py|) adapts the data to the format required by DualGraphRAG. 
The resulting processed dataset is organized as a set of separate directories, one per original HTML file. Each directory contains the corresponding HTML file, a markdown file storing the extracted textual content, a JSON file for the extracted structured data and a metadata JSON file with \verb|file-| prefix. 
This latter file stores the metadata associated with the original HTML file and lists the derived files available for that page. In particular, section \verb|content| points to the extracted text of the page, stored in the markdown file, while section \verb|prescience| points to the JSON file containing the extracted structural data. Either section may be absent if the corresponding data type is unavailable. 

The \emph{structured} part of the data (JSON files) collects the specifications of all products the specific HTML page describes, as separate objects. The information is taken from the \verb|specs.json| file, with simple processing ensuring a consistent structure of the object, parsing the structured attributes where necessary. The name, product range name and list of categories are individual top-level fields, while any other entries are converted into additional rows in a single dictionary collecting the available specifications. Each row in the dictionary has a key consisting of table section name and entry label separated by a comma. This consistent structure makes importing the data into any database easier.

The \emph{textual} content (markdown files) is extracted heuristically from the HTML, ignoring scripts, CSS and content-less structural elements, which account for the majority of the original file's size. Extraction focuses on the sections that contain useful information in the form of product or feature descriptions, FAQ entries, etc. The remaining text is mostly irrelevant to the actual products, often repeats on many or all pages, and includes for example various disclaimers, additional menus, financing options. On average only about 20\% of visible text and meaningful metadata (like title or alt attributes) on each page is relevant (see Table \ref{tab:text_extraction_stats}). Some predefined headers, e.g. containing the section of the page a given portion of text is from, are also inserted in the markdown. The structured information, already stripped of formatting and other noise 
is also attached in a plain text form to the markdown file, giving text-only based systems (e.g. TKG, Microsoft GraphRAG, RAPTOR, etc.) access to all available information.

\subsection{SKG: Datalog Preprocessing Rules} \label{appendix:datalog}
The manually defined rules shown in Listing~\ref{lst:datalog} are applied to the SKG as a means of making certain useful features easily accessible and adjusting data structure. The purpose of each line is shortly explained in the comments.

\newtcolorbox{promptbox}{
  fontupper=\footnotesize\ttfamily,
  left=4pt,right=4pt,top=0pt,bottom=0pt,
  arc=0mm,
}
\begin{listing*}
\begin{promptbox}
\begin{lstlisting}[breakindent=0pt,breaklines,basicstyle=\scriptsize\ttfamily,caption=Default Datalog rules, label=lst:datalog,captionpos=b]
% Common type for SKG nodes
[?s, a, skgt:SKG_Entity] :- [?s, a, ?c], FILTER(?c != skgt:UTKG_Entity) .
   
% skg:hasFeature maps to Entry in table if entry has value "yes"
[?p, skg:hasFeature, ?f], [?f, rdf:type, skgt:Feature] :- 
[?p, skg:hasSpec, ?s], [?s, skg:inEntry, ?f], [?s, skg:hasValue, skg:yes] .

% skg:hasFeature maps to skg:8k_recording_support (created if not exists)
% if some feature ID in skg:video_recording_resolution contains "8k"
[?p, skg:hasFeature, skg:8k_recording_support],
[skg:8k_recording_support, rdf:type, skgt:Feature],
[skg:8k_recording_support, skg:hasName, "8K Recording Support"] :- [?p, skg:hasSpec, ?s], 
[?s, skg:inEntry, skg:video_recording_resolution], [?s, skg:hasValue, ?f],
FILTER(REGEX(str(?f), "8k")) .

% skg:hasFeature maps to skg:5g_support (created if not exists)
% if at least one of the specific 5G features is present
[?p, skg:hasFeature, skg:5g_support], [skg:5g_support, rdf:type, skgt:Feature], 
[skg:5g_support, skg:hasName, "5G Support"] :- [?p, skg:hasSpec, ?s], [?s, skg:hasValue, ?f5g], 
FILTER (?f5g IN (skg:5g_sub6_fdd, skg:5g_sub6_tdd, skg:5g_sub6_sdl ) ) .

% skg:hasFeature maps to skg:4g_support (created if not exists)
% if at least one of the specific 4G features is present
[?p, skg:hasFeature, skg:4g_support], [skg:4g_support, rdf:type, skgt:Feature], 
[skg:4g_support, skg:hasName, "4G Support"] :- [?p, skg:hasSpec, ?s], [?s, skg:hasValue, ?f4g], 
FILTER (?f4g IN (skg:4g_lte_fdd, skg:4g_lte_tdd ) ) .

% Price is a special property and should be easily found
[?product, skg:hasPrice, ?price] :- [?product, skg:hasSpec, ?spec], 
[?spec, skg:inEntry, ?entry], [?entry, skg:hasName, "Price"], [?spec, skg:hasValue, ?price] .

% Products also belong in categories
[?p, skg:belongs, ?c] :- [?p, skg:variantOf, ?pr], [?pr, skg:belongs, ?c] .
\end{lstlisting}
\end{promptbox}
\end{listing*}

\subsection{SKG: Schema}\label{appendix:schema}
\begin{figure*}[h]
  \centering
  \includegraphics[width=\linewidth]{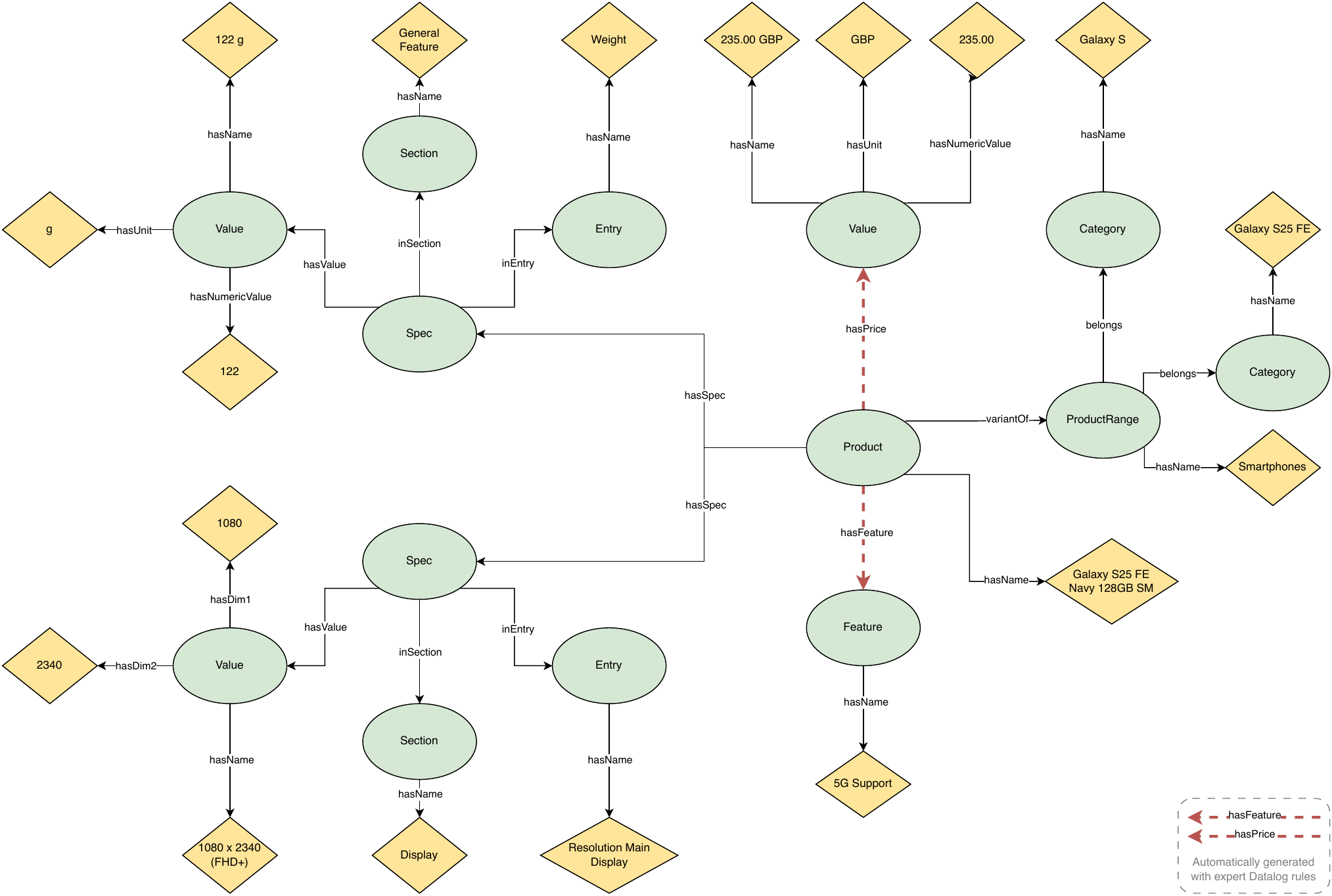}
\caption{Schema of the Symbolic Knowledge Graph (SKG) used in DualGraph.}  
\label{fig:schema}
\end{figure*}

The SKG's schema is presented in Figure \ref{fig:schema}, with the individual entity types described in Table \ref{tab:schema_classes} and predicates listed in Table \ref{tab:schema_predicates}. This schema is sufficient to describe the imported data, however, to facilitate SPARQL generation and higher level understanding,  additional custom types and predicates can be added with Datalog rules (see previous section).

\begin{figure*}[h]
  \centering
  \includegraphics[width=0.8\linewidth]{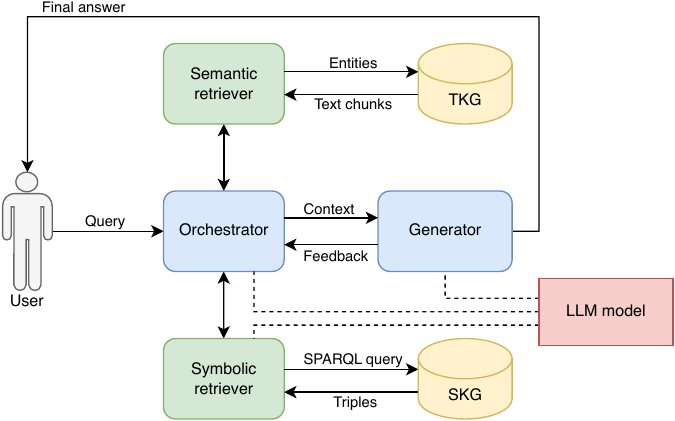}
\caption{Overview of the DualGraph querying pipeline. Blue components denote operation modules, green components denote retrieval functions, yellow components denote data storage, and LLM is depicted in red.}  \label{fig:query_diagram}
\end{figure*}

\begin{table*}
  \caption{Description of classes used in the schema.}
  \label{tab:schema_classes}
  \footnotesize
      \centering
  \begin{tabular}{lp{0.6\linewidth}}
    \toprule
    Class & Description \\
    \midrule
    skgt:Category & Wide category of products, e.g. skg:smartphones, skg:projectors or skg:galaxy\_z, extracted from URL \\
    skgt:ProductRange and skgt:Product & Specific range of products or individual product, e.g. Galaxy S22 5G or Galaxy S25 FE. Product is used for differentiating variants, e.g. each color version is an individual product. Where product has no variants there is one node with both types \\
    skgt:Spec & Single specification, an auxiliary node for representing given entry in specification table, it is connected to nodes of type Product. The only type without a name (skg:hasName) \\
    skgt:Section & Always connected to specification node, represents name of specification table section, directly extracted from the header, e.g. skg:storage\_memory, skg:processor, skg:camera. Sectionless entries are grouped in an artificial default section "Specifications" \\
    skgt:Entry & Always connected to specification node, represents name of specification table entry, e.g. skg:multi\_view, skg:mount, skg:battery\_type \\
    skgt:Value & Always connected to specification node, represents value in given specification table section and entry, e.g. skg:40\_w, skg:glare\_free, skg:included. The name (skg:hasName) is the literal string from the page, or a fragment of it resulting from splitting on commas into multiple entries where necessary \\
    skgt:Feature & Additional features of a product not directly present in the specification table but deduced from it using expert-provided Datalog rules, e.g. skg:5g\_support, skg:8k\_recording\_support. Connected to Product, but not connected to Spec as it is often deduced from multiple rows \\
    skgt:UTKG\_Entity & Entity extracted in TKG. May coincide with any other type (except Spec) if name matches exactly \\
    \bottomrule
  \end{tabular}
\end{table*}

\begin{table*}
  \caption{Description of predicates used in the schema.}
  \label{tab:schema_predicates}
    \footnotesize
    \centering
  \begin{tabular}{lp{0.6\linewidth}}
    \toprule
    Predicate & Description \\
    \midrule
    rdf:type & Alias for 'a', means assigning a given node to a type \\
    skg:hasName & Each node type (except for Spec) has a name in natural language, retrieved directly from the page \\
    skg:variantOf & Means that a given Product is a variant of given ProductRange. For products with no variants this relation is cyclical, as the same node is both ProductRange and Product \\
    skg:belongs & Means that a given ProductRange is in a given Category \\
    skg:hasDescription & UTKG\_Entities have longer, a LLM-provided description in natural language \\
    skg:hasSpec & Connects nodes of type skgt:Product and skgt:Spec \\
    skg:inSection & Connects nodes of type skgt:Spec and skgt:Section \\
    skg:inEntry & Connects nodes of type skgt:Spec and skgt:Entry \\
    skg:hasValue & Connects nodes of type skgt:Spec and skgt:Value \\
    skg:hasFeature & Connects nodes of type skgt:Product and skgt:Feature \\
    skg:hasNumericValue & If a Value could be meaningfully parsed as a number, links the Spec to a numeric literal containing the result \\
    skg:hasDim\# & If a Value specifies dimensions (2D or 3D) these point to individual numeric literals for each dimension \\
    skg:hasUnit & If a unit could be deduced or parsed out of the Value (or potentially Entry – not currently done), this relation points to a string literal specifying the unit. Units are normalized, eg. 'l', 'L' all point to 'l' \\
    \bottomrule
  \end{tabular}
\end{table*}

\subsection{UnWeaver TKG}
\label{sec:unweaver}
\begin{figure*}[h]
  \centering
  \includegraphics[width=\linewidth]{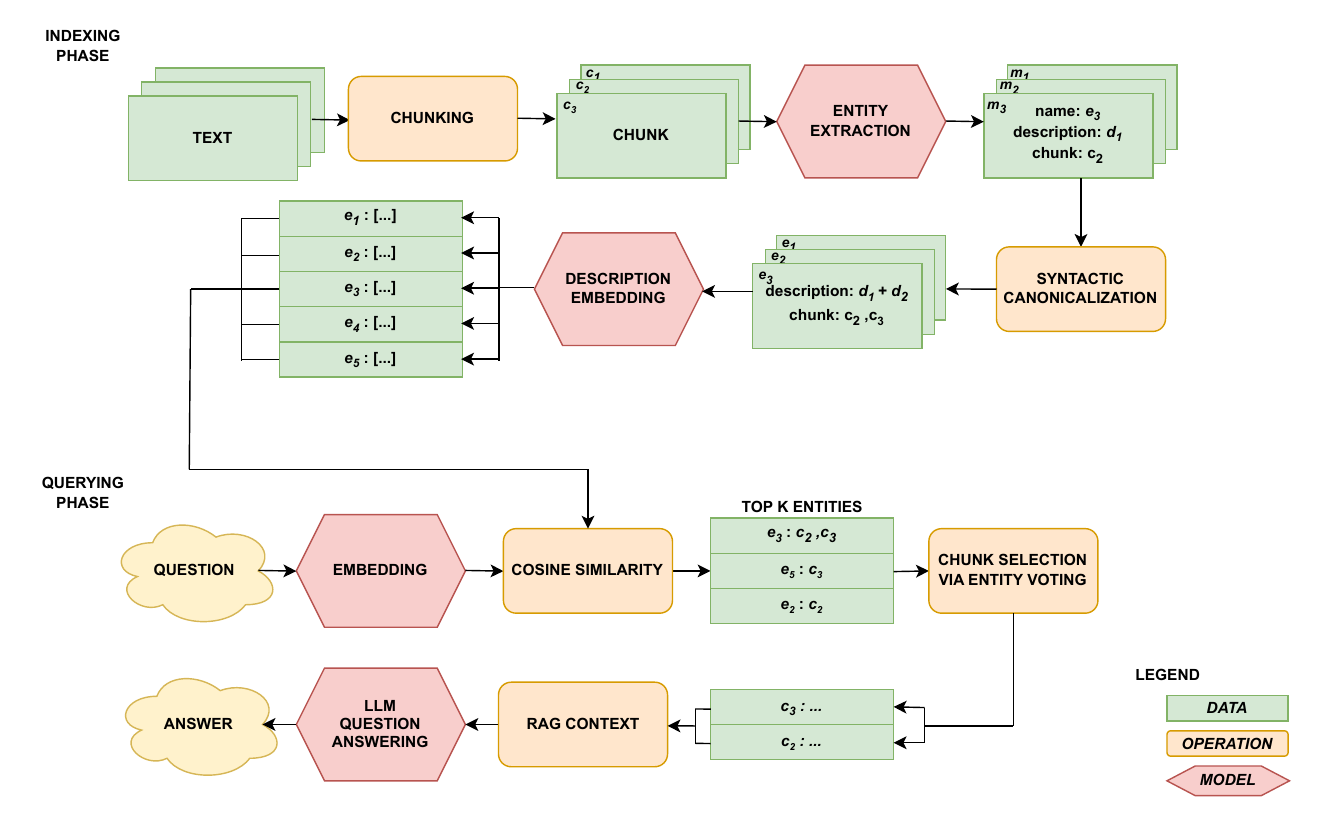}
  \caption{\label{fig:unweaver}Architecture of UnWeaver (Figure 1 from \cite{tuora2026unweavingknotsgraphrag})}
  
\end{figure*}

We generate the TKG as in UnWeaver \cite{tuora2026unweavingknotsgraphrag}, as it is conceptually simple and efficient in terms of token usage. UnWeaver uses an LLM to extract a list of untyped entities out of each text chunk. Relations are not extracted, as this additional layer contributes little to the actual end-to-end QA performance. Each entity is fully characterized by its LLM-generated name, and a short description. 
After the extraction, all the entities are merged by name to reduce cross-chunk information redundancy, which yields canonicalized entities, with aggregate descriptions. The final result is a bipartite graph index linking entities with chunks in which they are mentioned. The full architecture of UnWeaver can be seen in Figure \ref{fig:unweaver}. In the query phase entities (represented by their descriptions) are used as an intermediate layer for the purpose of identifying question-relevant chunks, with no LLM-generated content being shown in the context provided to the LLM for generating the final answer.

\subsection{Refinement Details}
After constructing both the SKG and the TKG, we align their nodes using normalized name matching (stemming). This relies on a simple heuristic where we cast all the names to lower cases, remove any leading or trailing whitespace as well as any prepositions like "a" or "the" (stop words removal). Then we align them using exact name matching.

We then apply an additional refinement step to improve alignment quality by using the aligned entities from previous step as seed to train a self supervised contrastive aligner. For that we calculate the embeddings of SKG by finding the largest connected component and running node2vec on it to obtain the embeddings. For TKG we rely on dense textual embeddings from a pre-trained embedder such as Qwen3-Embedding-4B. Then we take batches from common nodes of SKG and TKG and minimize the information noise-contrastive estimation (InfoNCE) loss to train the aligner. Once the aligner is trained we apply it on the non-common nodes of SKG and TKG to obtain the most likely candidates for alignment. Those candidates are then evaluated in two independent passes by LLM judge to determine if the two nodes should be aligned. The unfortunate limitation of this method is that LLM judge can only make the decision based on the names of proposed nodes since for the SKG nodes we can not offer a textual description of the node and so to keep the selection process symmetric we chose not to offer the description for the TKG nodes either.

\subsection{LLM as a Judge}
The \textbf{LaaJ} evaluation uses \texttt{gpt-oss-120b} as a judge to choose the better out of two competitors. Each pair of systems is tested across all examples in both permutations of placement. The final score of a system is the average winrate across all settings and all competitors.

\begin{figure}
    \centering
    \includegraphics[width=0.6\linewidth]{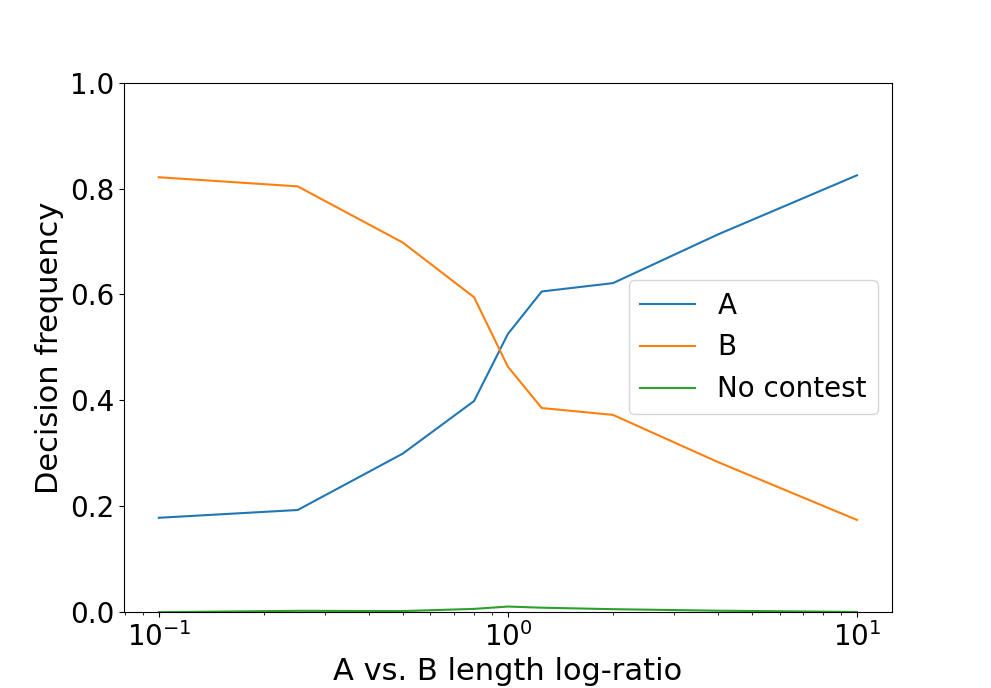}
    \caption{LaaJ decision frequency is dependent on A~vs~B answer length log-ratio. \texttt{No contest}  correspond to cases where the LLM response cannot be parsed as either A or B.}
    \label{fig:laaj_length_bias}
\end{figure}

The \textbf{LaaJ} shows a consistent preference for longer answers, as shown in \autoref{fig:laaj_length_bias}.

\section{Additional details about the data}

\subsection{Layout types}\label{appendix:layouts}
\begin{figure*}[t!]
    \centering
    \begin{subfigure}[t]{0.5\textwidth}
        \centering
        \frame{\includegraphics[width=0.9\linewidth]{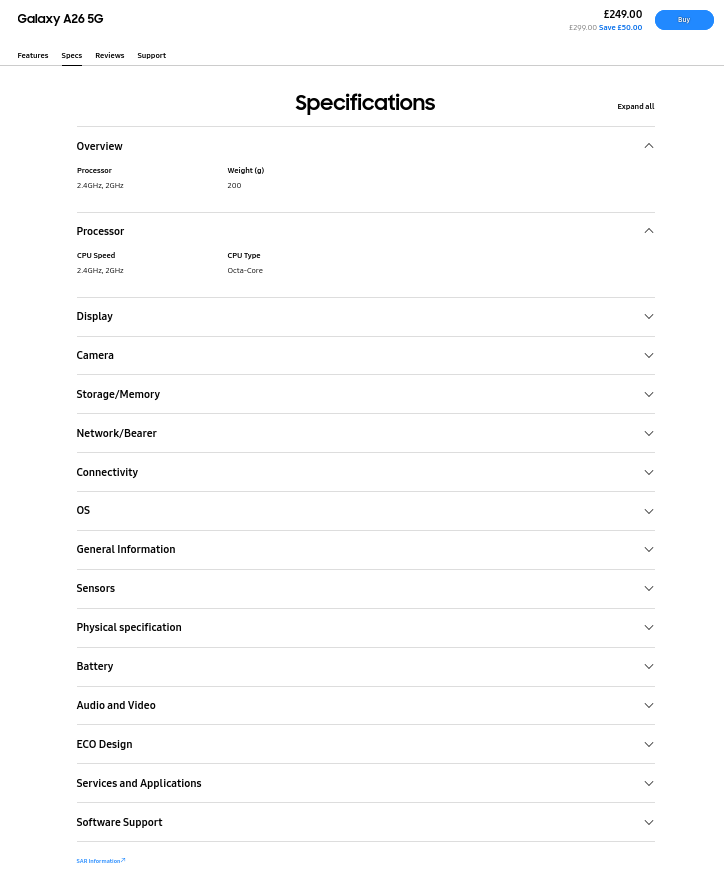}}
        \caption{Example of an A-layout}
    \end{subfigure}%
    ~ 
    \begin{subfigure}[t]{0.5\textwidth}
        \centering
        \frame{\includegraphics[width=0.9\linewidth]{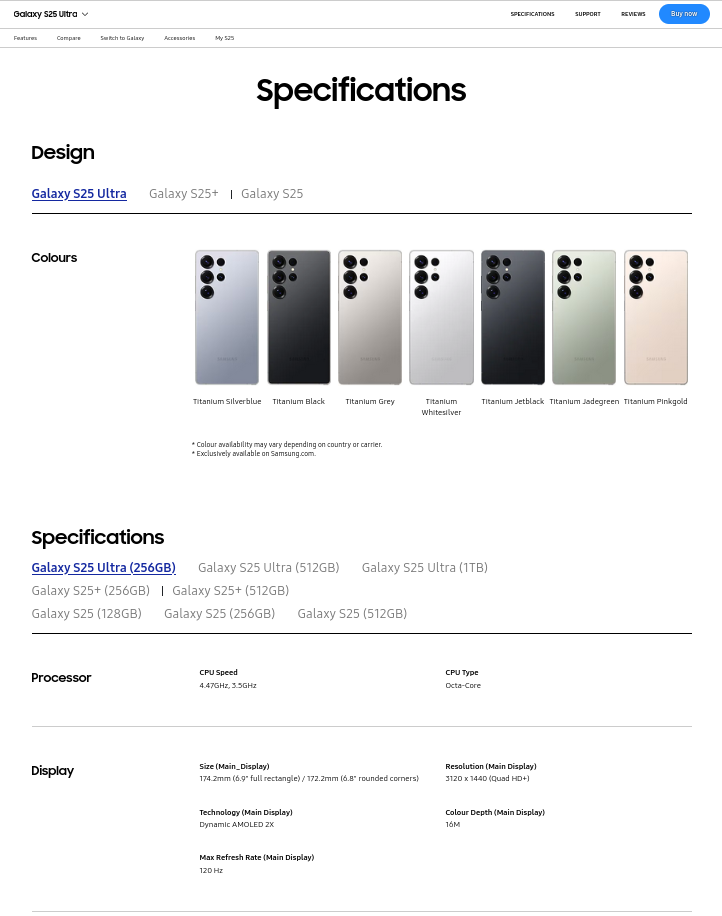}}
        \caption{Example of a B-layout}
    \end{subfigure}
    \vspace{0.5cm}

    \begin{subfigure}[t]{0.5\textwidth}
        \centering
        \frame{\includegraphics[width=0.9\linewidth]{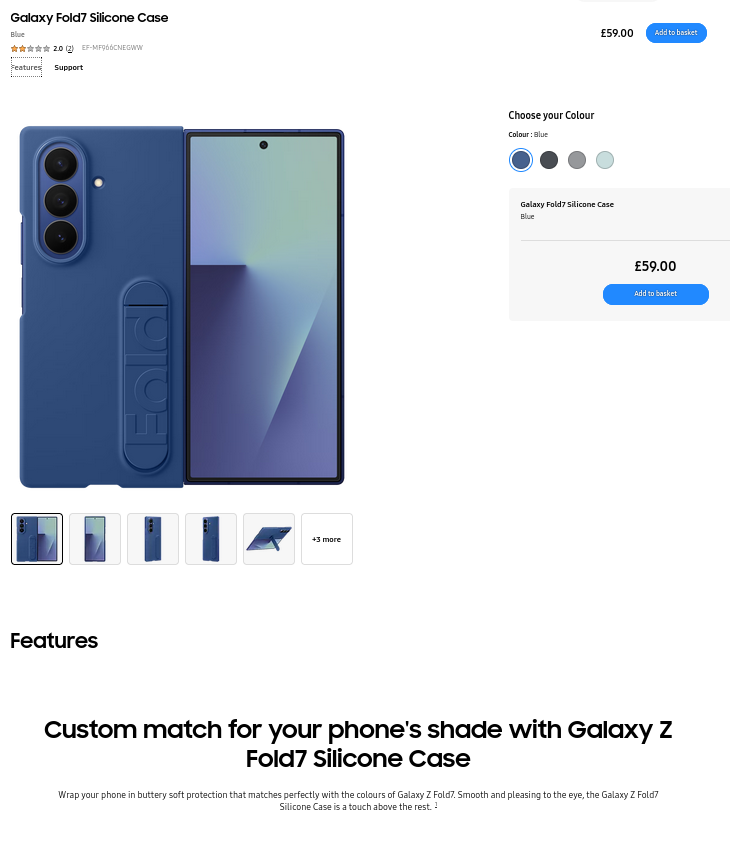}}
    \caption{Example of a C-layout}
        \end{subfigure}%
    \caption{Specification page layout examples}
        \label{fig:layouts}
\end{figure*}

The product pages conform to three rough layouts, which we label A, B, and C (see Fig~\ref{fig:layouts}). The A type is the most general one, occurring across categories, whereas B is reserved for premium smartphones, and C is mostly used for accessories. B layout contains rich multimedia content, but most importantly uses a separate webpage (as opposed to a tab) to list specifications, which we use to generate triples for injection into the knowledge graph. Moreover the pages containing specifications in the B layout do not conform to the $1:1$ mapping between products and pages, instead listing multiple variants, and even multiple product ranges. This requires dedicated scraping logic: dynamically loading all the possible combinations of parameters on the specification page, to obtain all the information. For C layout there is no table containing specifications.

It should also be observed that some information (most importantly the price) is often not present on the product page itself, but instead must be carried from elsewhere (the product list).

\subsection{Extracted Data}
The documents obtained from the shop page are multimodal, and a full representation of their content would have to include visual, and video processing. In this work we focus only on textual information, and structured information contained in the specification. In \autoref{tab:text_extraction_stats} we present the quantitative statistics about the length of the raw documents, their extracted textual content, and final markdown files (without the structural parts contained in the tables). It is clear that the pure markdown text, which serves as the basis of TKG creation is rather limited, and most information is presented using other means (including the specification tables).

\begin{table*}
  \caption{Textual information in HTML data -- statistics for the full dataset.}
  \label{tab:text_extraction_stats}
  \centering
  \resizebox{0.6\linewidth}{!}{
  \begin{tabular}{lcc}
    \toprule
    Processing phase & Avg. character count & Avg. token count \\
    & (min/max) & (min/max) \\
    \midrule
    Raw HTML & 1,600,592 & 650,714 \\
    & (1,248,179/2,728,986) & (557,184/991,785) \\
    Textual content & 48,735 & 10,691 \\
    & (21,599/104,381) & (5,124/21,344)\\
    Final extracted MD & 8,101 & 2,035 \\
    & (364/43,154) & (108/10,090) \\
    \bottomrule
  \end{tabular}
  }
\end{table*}

\subsection{Question Generation}
The process of question generation consisted of several steps:
1) Feature Identification: We first identified key smartphone attributes commonly considered in purchase decisions. Prior research and market analyses highlight features such as price, brand, network connectivity (e.g., 5G capability), battery capacity, display technology, camera quality, memory, and storage as critical factors for consumers. These attributes formed the foundation for our question criteria.
2) Constraint Definition: For each selected attribute, we defined concrete constraint values or categories to ensure measurability. Numeric features were given threshold values (e.g., 'under GBP 300' for price, '>4000 mAh' for battery capacity) while categorical features were specified by exact values (e.g. 'AMOLED display', '5G connectivity = Yes'). The constraint levels were chosen to reflect meaningful consumer thresholds: for instance, price caps correspond to common budget segments (budget vs. mid-range) and battery capacity cutoffs align with above-average battery life expectations. 
3) Query Formulation: We then composed a set of evaluation questions. Each query explicitly enumerates all conditions to avoid ambiguity. For example, one formulated question asks for 'Samsung smartphones under GBP 300 that support 5G, have at least a 4000 mAh battery, and feature an AMOLED display.' This multi-faceted structure ensures the question is highly specific---only devices meeting all stated criteria qualify as answers. Such questions effectively mimic a faceted search query a consumer might use on an e-commerce site, filtering by brand, price range, and specific features. The specificity of the constraints makes the expected results clear-cut and reproducible: any researcher applying the same filters to the same dataset should retrieve an identical set of smartphones.

\subsection{More about question categories}\label{sub:questionTypes}

To better illustrate the diversity of the benchmark and the complementary strengths of the SKG and TKG, we provide several representative examples below.

\textbf{Inverse Queries} require retrieving all entities satisfying a given property or relation. Instances of these are: "Which Samsung phones support wireless charging?"
"Which Galaxy models feature a 200MP wide-angle camera?"
"Which Samsung phones support expandable storage via microSD card?"

These questions primarily involve structured attribute lookup and entity filtering over relatively stable specifications. More nuanced inverse queries of these type are e.g.,: "Which Samsung phones currently have a removable battery?"
"Which Samsung phones are considered water-resistant or waterproof?"
"Which phones are marketed as rugged or MIL-STD compliant?"

\textbf{Multi-Condition Queries} combine several constraints simultaneously, often requiring compositional retrieval. Examples of these queries include: "Which Samsung phones under GBP 500 offer a 120Hz display, 5G, and at least a 5000mAh battery?"
"What phones come with 8GB RAM, 256GB storage, and cost under GBP 600?"
"Which Samsung phones support Samsung DeX, have 5G, and are under GBP 1000?"

Some constraints are implicit, fuzzy, or context-dependent: "Which Samsung phones are durable enough for outdoor work and also support dual SIM?"
"Which foldable Samsung phones are still practical for everyday use and long battery life?". 

Here, the system must combine structured specifications with textual or contextual knowledge distributed across documentation.

\textbf{Group Comparison Queries} require aggregating and contrasting properties across multiple device families or products, as seen in the following examples: "How do the Galaxy A series and Galaxy S series differ?"
"Compare the features of Galaxy Z Fold vs. Galaxy Z Flip phones."
"What changed from Galaxy Z Flip6 to Galaxy Z Flip7?"

Answers to these typically require synthesizing multiple heterogeneous relations into coherent comparative narratives rather than returning isolated facts. Sometimes, the comparison is grounded in explicit specifications such as battery size, camera resolution, RAM, or display dimensions.

\textbf{Reasoning Queries} are the most open-ended category and require combining factual retrieval with user-oriented inference, e.g., "What's a good phone for someone who only uses basic apps like YouTube, Tinder, and WhatsApp?"
"Recommend a compact Samsung phone for easy one-handed use."
"Which Samsung phone is best for multitasking and productivity?"
"What Samsung phone would you suggest for someone who values a great display for movies?"

These questions often cannot be answered via direct attribute matching alone. Instead, they require synthesizing multiple signals (display quality, battery life, ergonomics, portability, productivity features, etc.) into a recommendation aligned with user intent.

\section{Full set of results}\label{appendix:full_results}

\begin{figure*}[t!]
    \centering
    \begin{subfigure}[t]{0.5\textwidth}
        \centering
        \includegraphics[width=\linewidth]{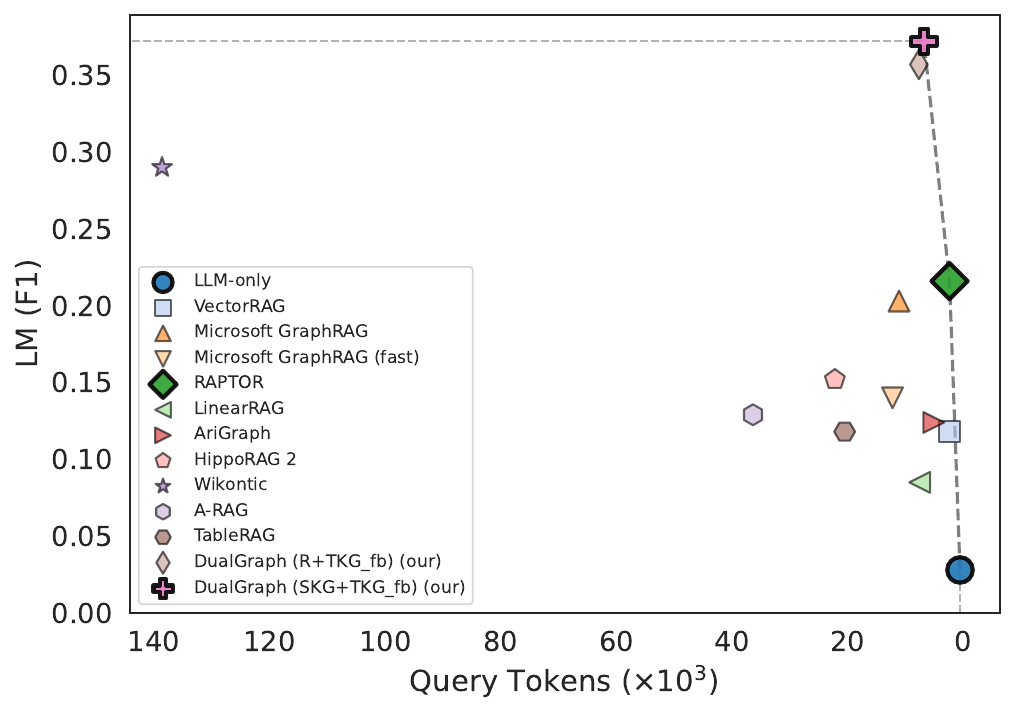}
        \caption{Querying token count.}
    \end{subfigure}%
    ~
    \begin{subfigure}[t]{0.5\textwidth}
        \centering
        \includegraphics[width=\linewidth]{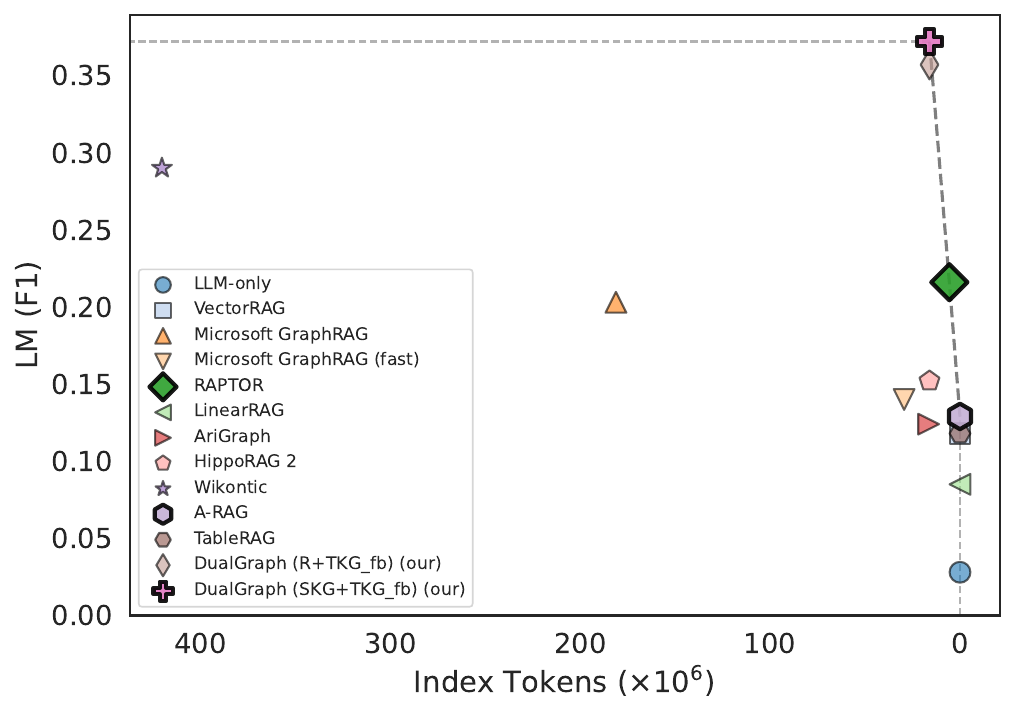}
        \caption{Indexing token count.}
    \end{subfigure}%
      \caption{Pareto-front comparison between DualGraph and state-of-the-art baselines in terms of list matching (F1) and querying/indexing token usage.}
    \label{tab:pareto_f1}
\end{figure*}

\begin{figure*}[t!]
    \centering
    \begin{subfigure}[t]{0.5\textwidth}
        \centering
  \includegraphics[width=\linewidth]{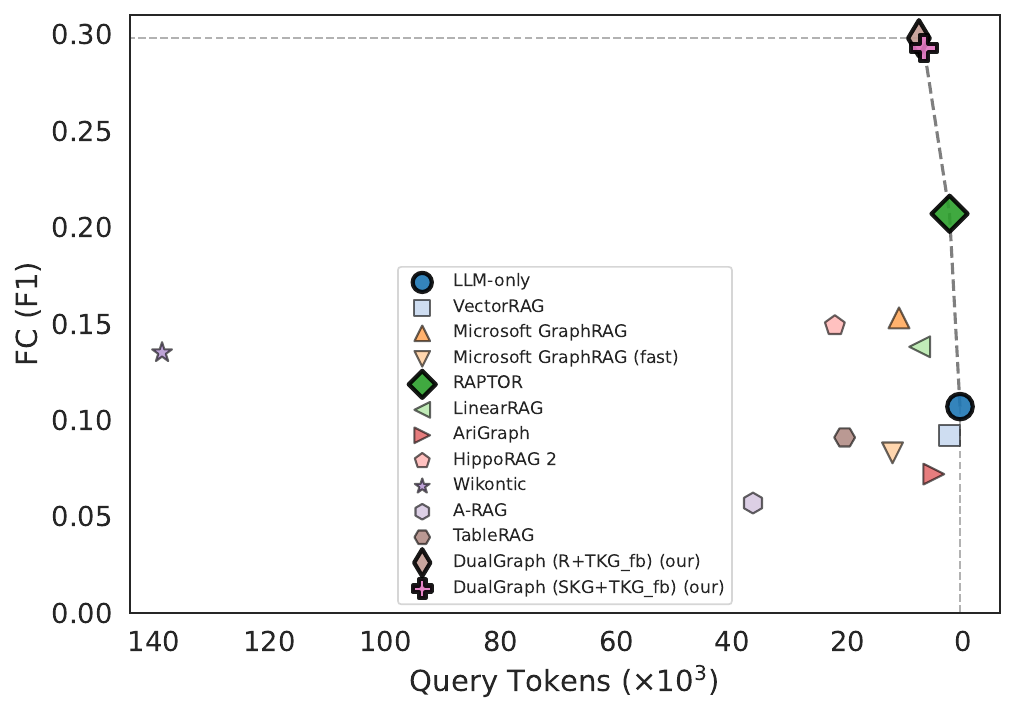}
        \caption{Querying token count.}
    \end{subfigure}%
    ~
    \begin{subfigure}[t]{0.5\textwidth}
        \centering
  \includegraphics[width=\linewidth]{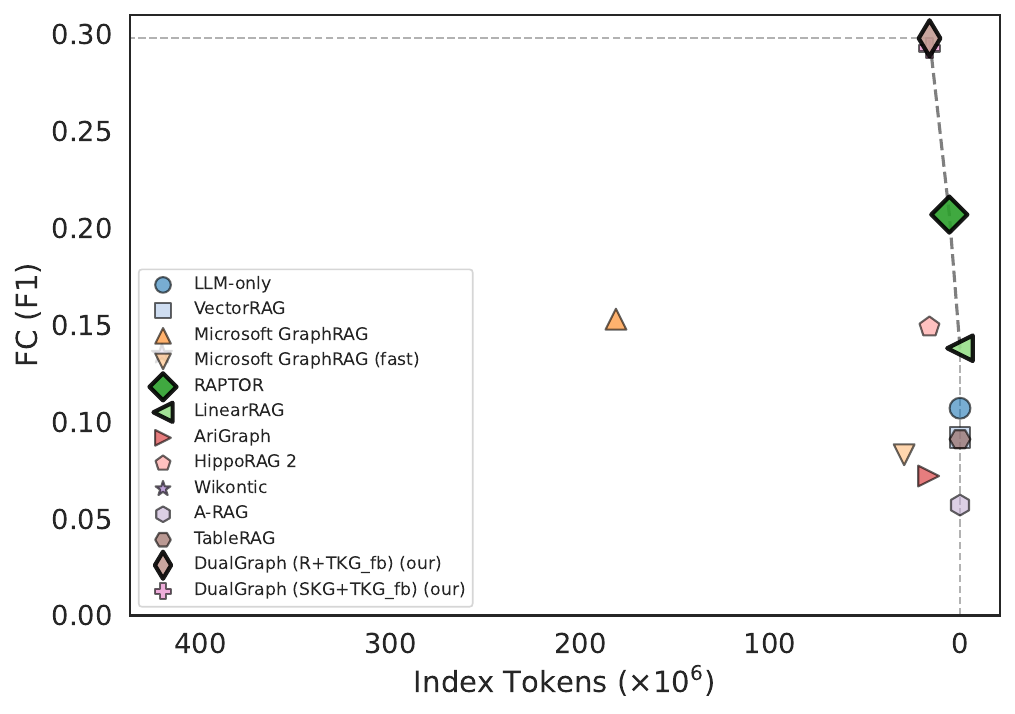}
          \caption{Indexing token count.}
    \end{subfigure}%

  \caption{Pareto-front comparison between DualGraph and state-of-the-art baselines in terms of factual correctness (F1) and querying/indexing token usage.}
      \label{tab:pareto_fc}

\end{figure*}

\begin{figure*}[t!]
    \centering
    \begin{subfigure}[t]{0.5\textwidth}
        \centering
  \includegraphics[width=\linewidth]{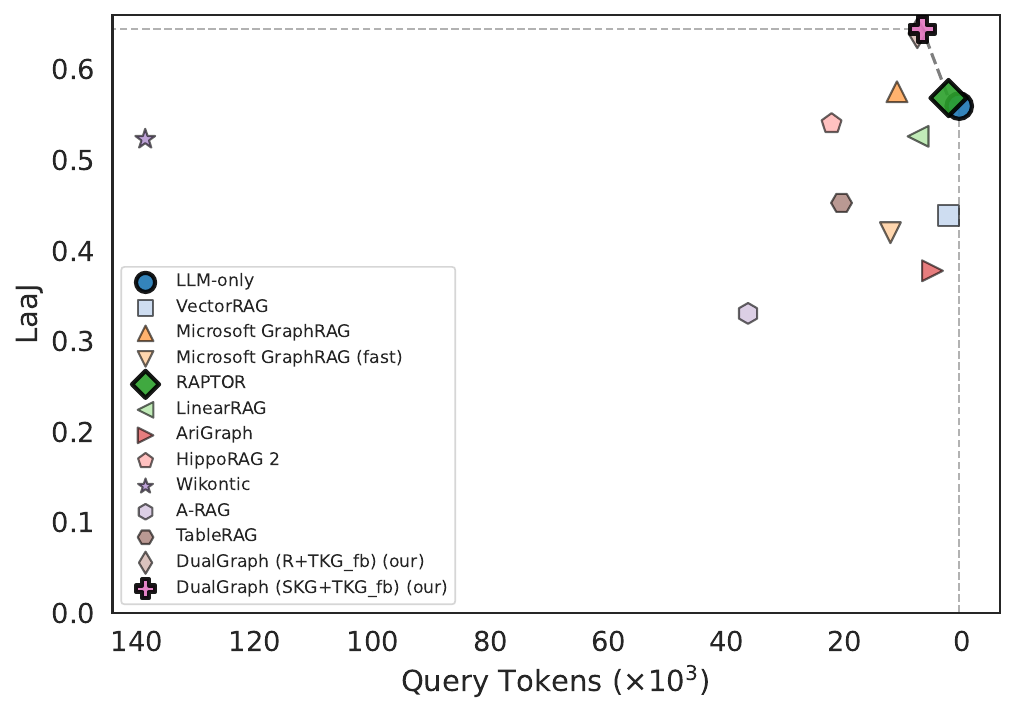}
  \caption{Querying token count.}
    \end{subfigure}%
    ~
    \begin{subfigure}[t]{0.5\textwidth}
        \centering
  \includegraphics[width=\linewidth]{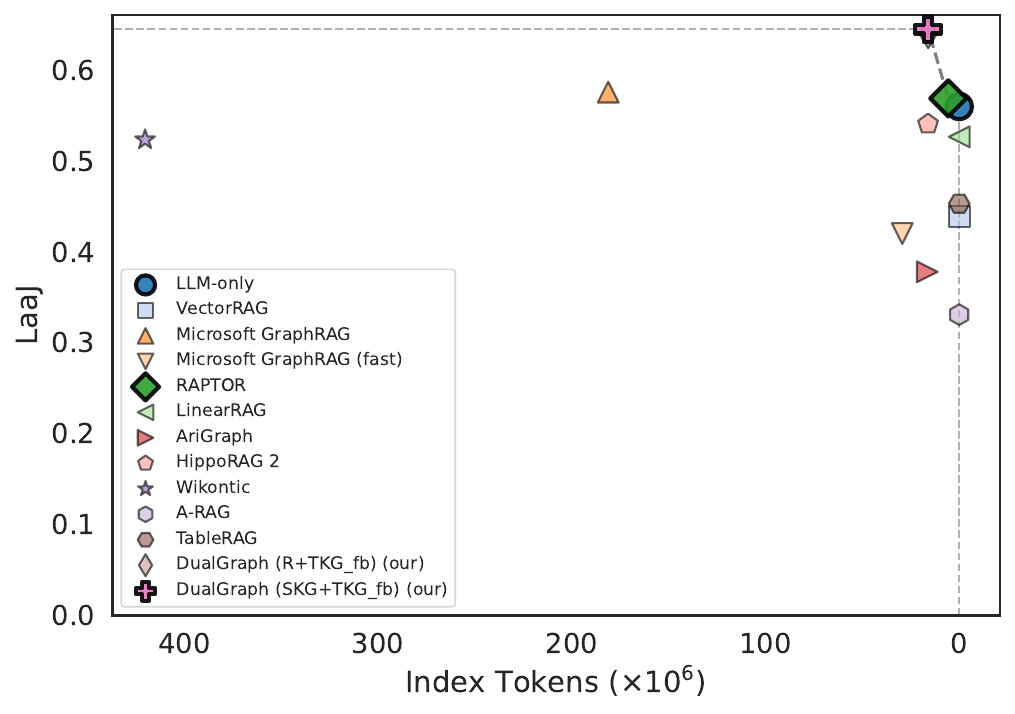}
   \caption{Indexing token count.}
    \end{subfigure}%
\caption{Pareto-front comparison between DualGraph and state-of-the-art baselines in terms of LLM-as-a-judge score and querying/indexing token usage.}
\label{tab:pareto_laaj}
\end{figure*}

\begin{figure*}[t!]
    \centering
    \begin{subfigure}[t]{0.5\textwidth}
        \centering
        \includegraphics[width=\linewidth]{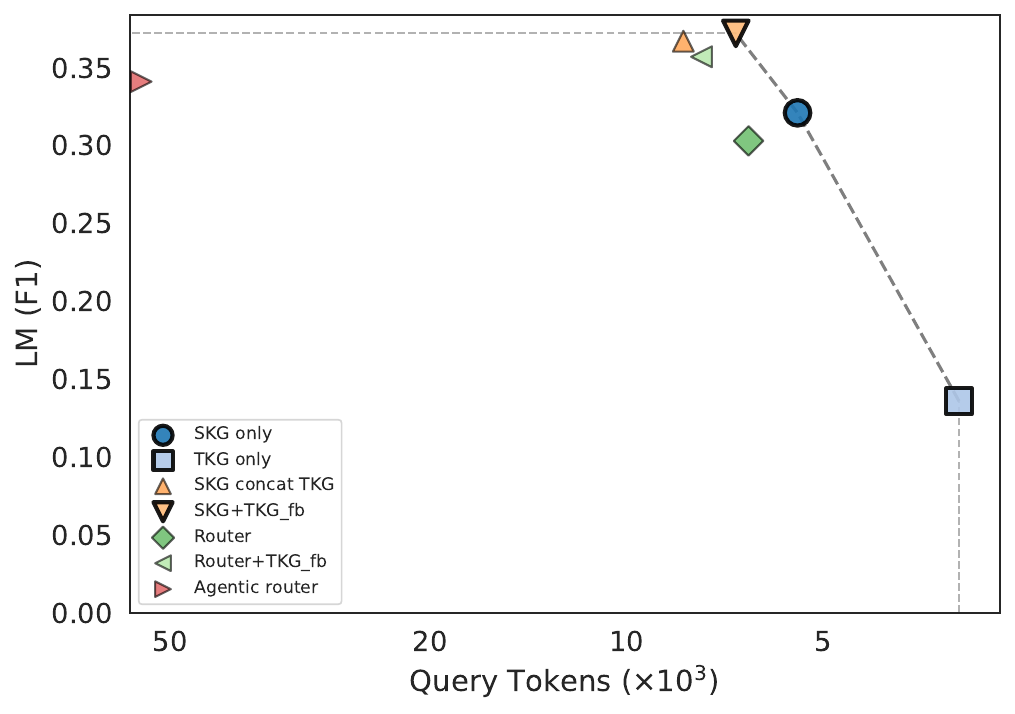}
        \caption{List matching F1 vs querying token count.}
    \end{subfigure}%
~
    \begin{subfigure}[t]{0.5\textwidth}
        \centering
  \includegraphics[width=\linewidth]{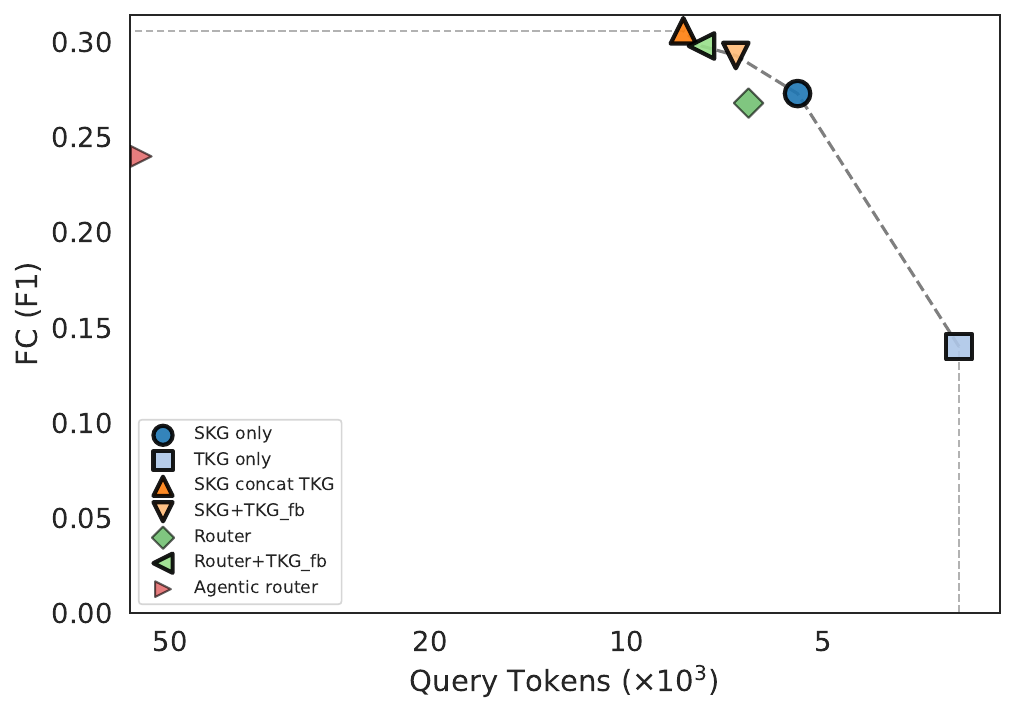}
  \caption{Factual correctness (F1) vs querying token count.}
\end{subfigure}%
    
    \begin{subfigure}[t]{0.5\textwidth}
        \centering
  \includegraphics[width=\linewidth]{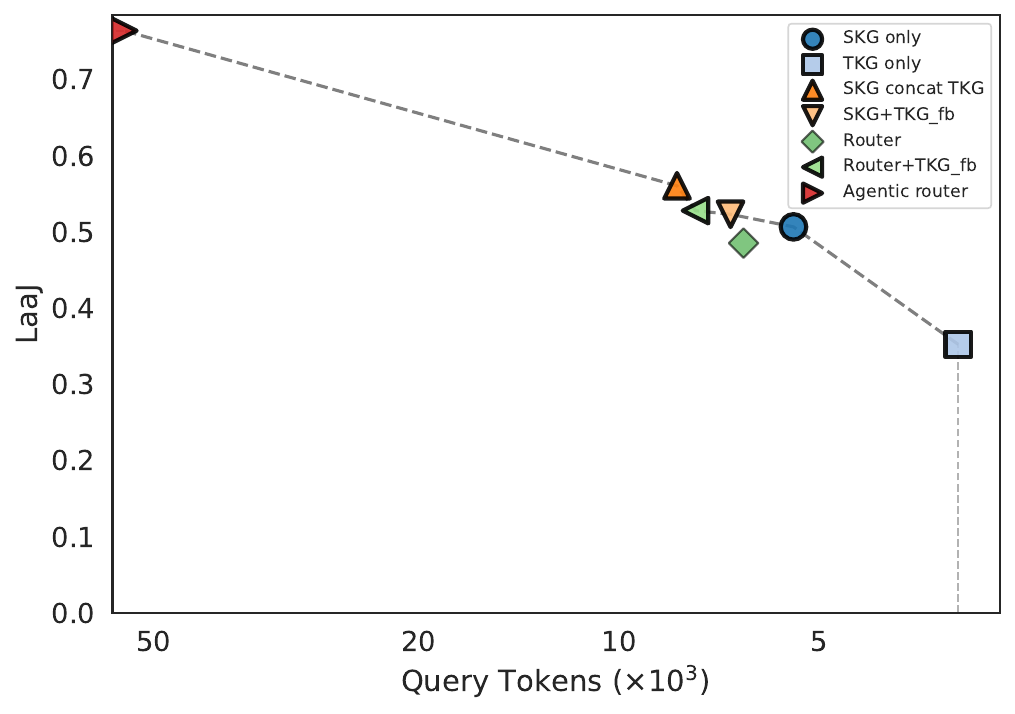}
  \caption{LLM-as-a-Judge score vs querying token count.}
\end{subfigure}%
\caption{Pareto-front comparison of different DualGraph variants across answer-quality metrics and query-time token usage.}
\label{fig:pareto_ablation}
\end{figure*}

\begin{figure*}[t!]
    \centering
    \begin{subfigure}[t]{\textwidth}
        \centering
  \includegraphics[width=\linewidth]{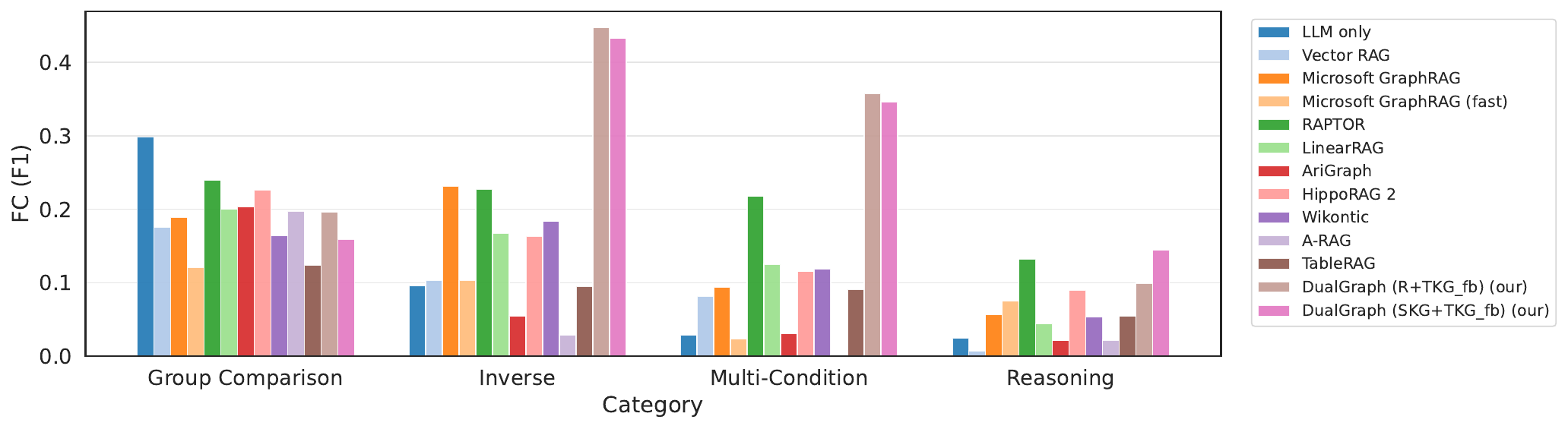}
    \end{subfigure}%
    
    \begin{subfigure}[t]{\textwidth}
        \centering
  \includegraphics[width=\linewidth]{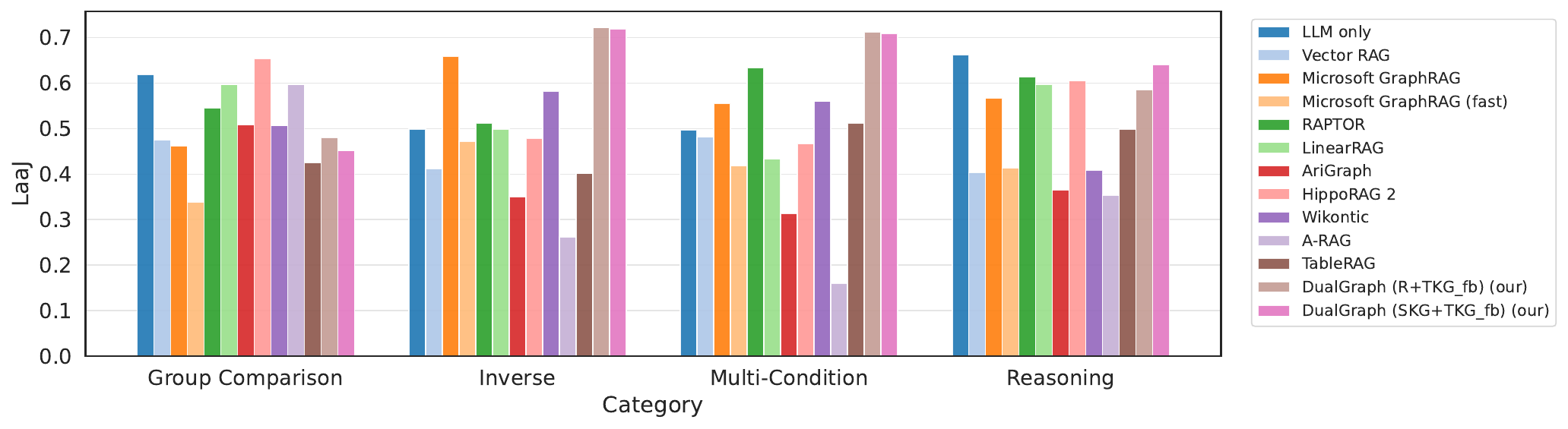}
\end{subfigure}

    \begin{subfigure}[t]{\textwidth}
        \centering
  \includegraphics[width=\linewidth]{list_f1_baseline.pdf}
\end{subfigure}
    \caption{Factual Correctness F1, List Matching F1, and LLM-as-a-judge scores by question category for DualGraph and state-of-the-art baselines.}
\label{fig:category_full}
\end{figure*}

\begin{figure*}[t!]
    \centering
    \begin{subfigure}[t]{\textwidth}
        \centering
     \includegraphics[width=\linewidth]{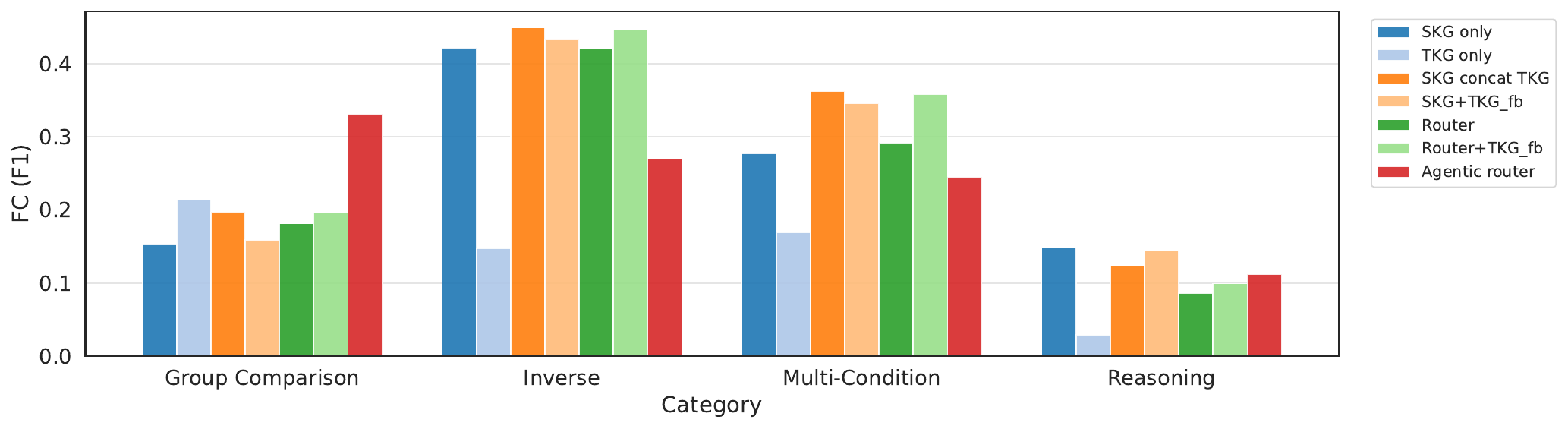}
    \end{subfigure}%
    
    \begin{subfigure}[t]{\textwidth}
        \centering
      \includegraphics[width=\linewidth]{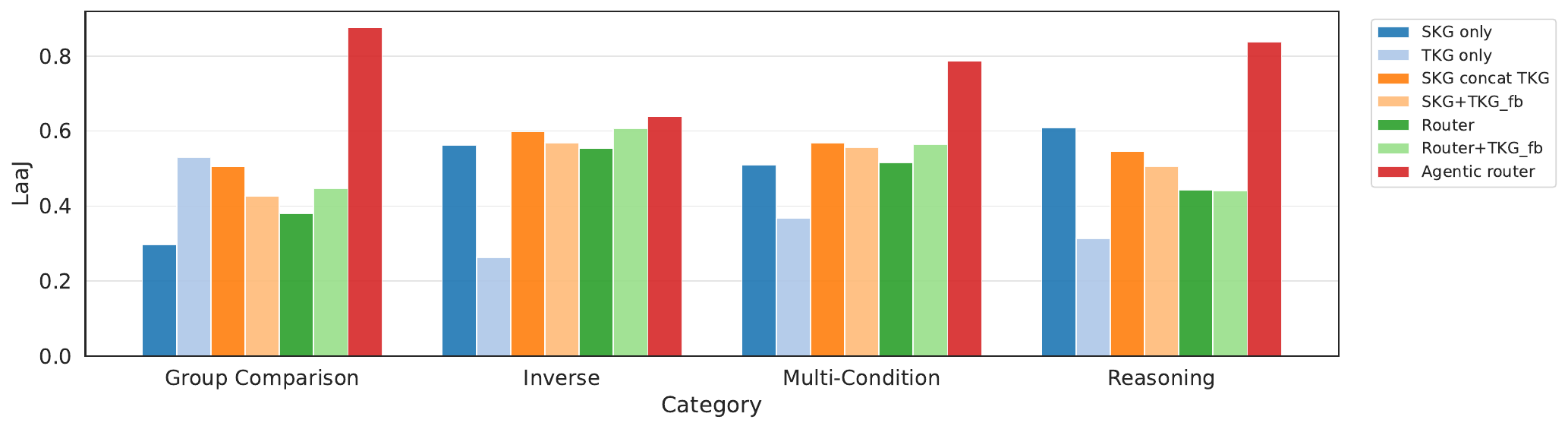}
    \end{subfigure}
    
    \begin{subfigure}[t]{\textwidth}
        \centering
     \includegraphics[width=\linewidth]{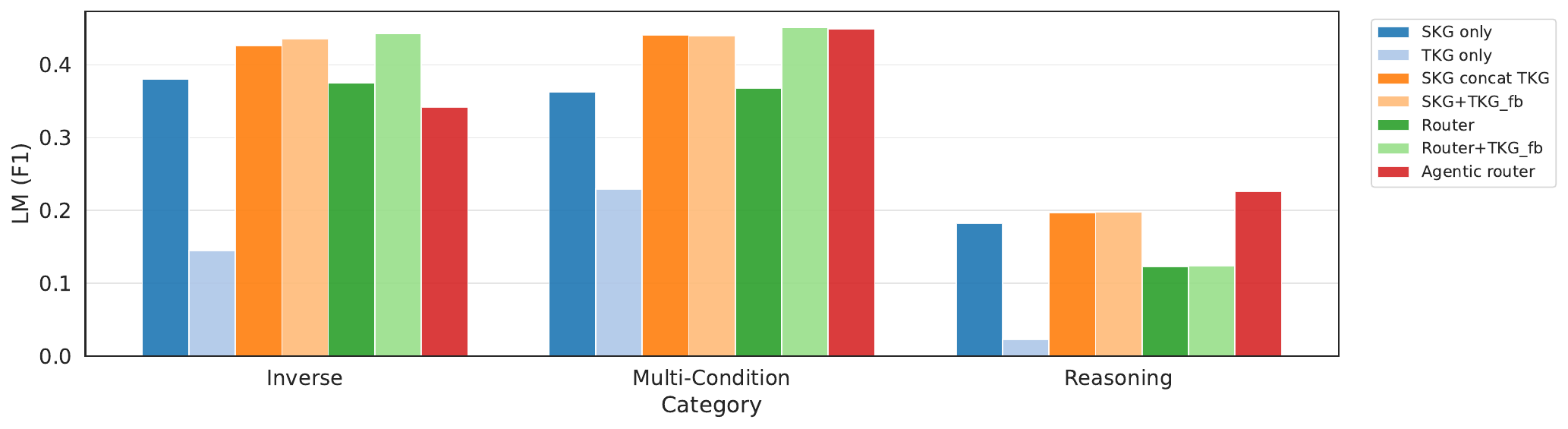}
    \end{subfigure}
    \caption{Factual Correctness F1, List Matching F1, and LLM-as-a-judge scores by question category for the ablation study of different DualGraph variants.}
\label{fig:category_ablation_full}
\end{figure*}

\begin{figure*}[h]
  \centering
  \includegraphics[width=\linewidth]{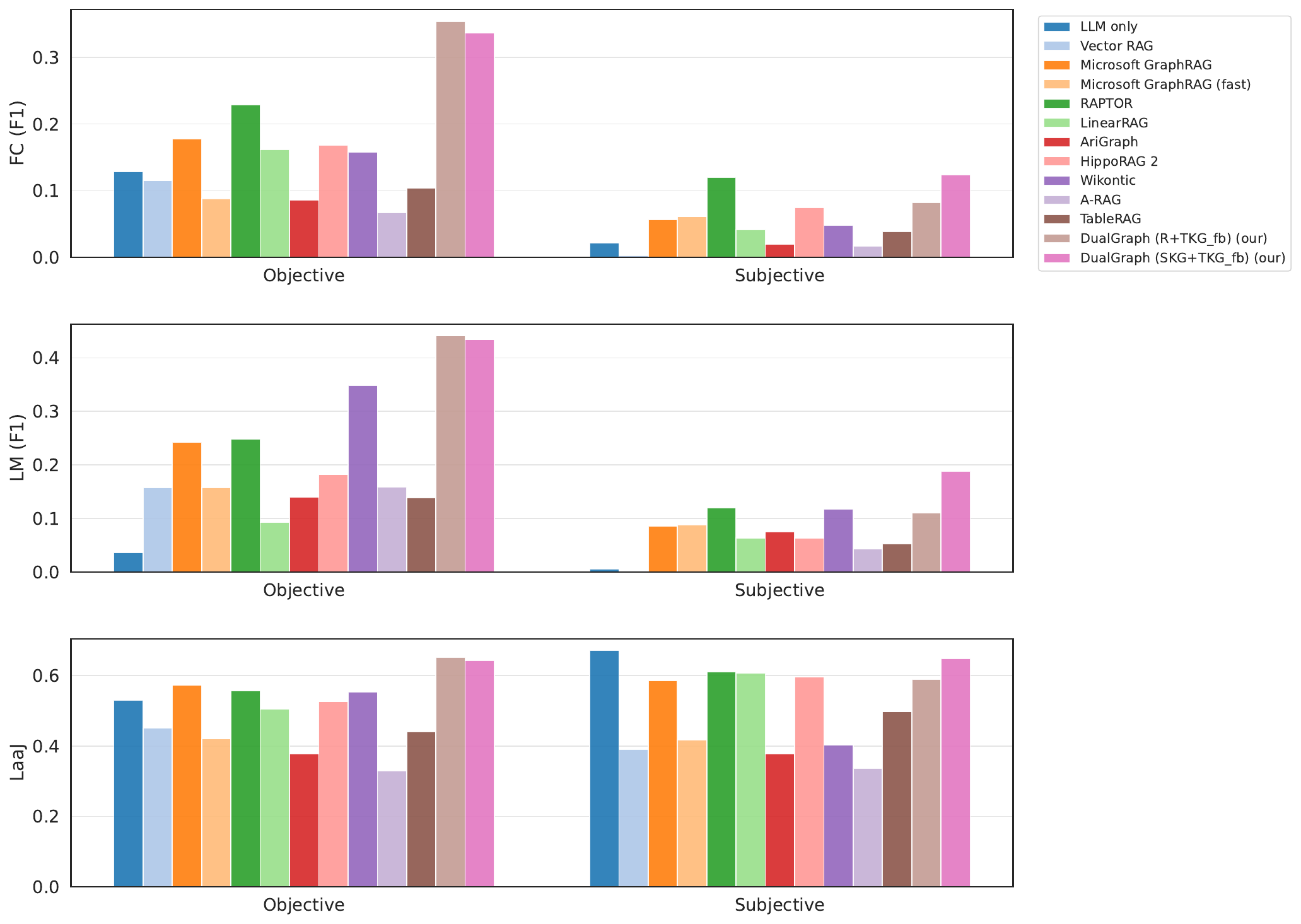}
\caption{Factual Correctness F1, List Matching F1, and LLM-as-a-judge scores separated by question type (objective vs.\ subjective).}
\label{fig:obj_sub}
\end{figure*}

This appendix reports the complete experimental results supporting the analysis in Section~\ref{sec:results}. Tables~\ref{tab:results_performance_sota_full}--\ref{tab:ablation_patterns_full} provide the full numerical results, while Figures~\ref{tab:pareto_f1}--\ref{fig:obj_sub} visualize the corresponding trends, trade-offs, and category-level breakdowns.

Table~\ref{tab:results_performance_sota_full} reports the full comparison between DualGraph and the state-of-the-art baselines in terms of factual correctness, list matching, and pairwise LLM-as-a-judge scores. 
Table~\ref{tab:results_usage_sota_full} reports the corresponding full results in terms of token usage, separating indexing and querying costs into prompt, completion, and total tokens. Together, these tables provide the complete version of the main comparison summarized in Table~\ref{tab:results_dualgraph_vs_sota}.

Table~\ref{tab:ablation_performance_full} reports the full ablation study over different DualGraph variants. It includes both answer-quality metrics and query-time token usage, complementing the discussion in Section~\ref{sec:ablation} (Table~\ref{tab:ablation_dualgraph_variants}). 
Table~\ref{tab:results_performance_objective_sota} separates results on objective and subjective questions, showing how performance changes when evaluation is restricted to questions with factual ground truth versus recommendation-style questions. 
Table~\ref{tab:ablation_patterns_full} reports the pattern ablation for both Router+TKG fallback and SKG+TKG fallback variants, measuring the contribution of Spec, Feature, Category, and Singular Node patterns to SKG retrieval and downstream answer quality.

Figures~\ref{tab:pareto_f1}--\ref{tab:pareto_laaj} show quality--cost trade-offs for the main comparison with baselines. Figure~\ref{tab:pareto_f1} plots list-matching F1 against querying and indexing token usage, Figure~\ref{tab:pareto_fc} shows the same analysis for factual correctness, and Figure~\ref{tab:pareto_laaj} reports the trade-off for LLM-as-a-judge scores. These figures illustrate the Pareto behavior of DualGraph relative to other RAG systems.

Figure~\ref{fig:pareto_ablation} provides the same quality--cost analysis between the different DualGraph variants, comparing list matching, factual correctness, and LLM-as-a-judge scores against query-time token usage. 
Figures~\ref{fig:category_full} and~\ref{fig:category_ablation_full} break down performance by question category: Figure~\ref{fig:category_full} compares DualGraph with the baselines, while Figure~\ref{fig:category_ablation_full} compares the internal DualGraph variants. Finally, Figure~\ref{fig:obj_sub} reports performance separately for objective and subjective questions, highlighting the additional difficulty of evaluating subjective recommendation-style queries.

\begin{table*}
\caption{Main results: comparison between DualGraph and state-of-the-art baselines in terms of answer quality.}  \label{tab:results_performance_sota_full}
  \resizebox{\linewidth}{!}{%
    \begin{tabular}{lccccccc}
      \toprule
       & \multicolumn{3}{c}{\textbf{Factual Correctness}} & \multicolumn{3}{c}{\textbf{List Matching}} & \textbf{LaaJ} \\
       \cmidrule[0.5pt](lr{0.3em}){2-4} 
       \cmidrule[0.5pt](lr{0.3em}){5-7} 
       & \multicolumn{1}{c}{\textbf{F1}} & \multicolumn{1}{c}{\textbf{Precision}} & \multicolumn{1}{c}{\textbf{Recall}} & \multicolumn{1}{c}{\textbf{F1}} & \multicolumn{1}{c}{\textbf{Precision}} & \multicolumn{1}{c}{\textbf{Recall}} & \multicolumn{1}{c}{} \\
      \midrule

      LLM only & 0.107 & 0.136 & 0.111 & 0.028 & 0.046 & 0.026 & 0.559 \\
      Vector RAG & 0.092 & 0.161 & 0.085 & 0.118 & 0.244 & 0.106 & 0.438 \\
      Microsoft GraphRAG \cite{edge2025localglobalgraphrag} & 0.153 & 0.240 & 0.152 & 0.203 & 0.375 & 0.188 & 0.575 \\
      Microsoft GraphRAG (fast) \cite{edge2025localglobalgraphrag} & 0.083 & 0.130 & 0.082 & 0.140 & 0.205 & 0.139 & 0.420 \\
      RAPTOR \cite{DBLP:conf/iclr/SarthiATKGM24} & 0.207 & 0.301 & 0.200 & 0.216 & 0.373 & 0.200 & 0.568 \\
      LinearRAG \cite{zhuang2025linearraglineargraphretrieval} & 0.138 & 0.197 & 0.163 & 0.085 & 0.172 & 0.079 & 0.526 \\
      AriGraph \cite{ijcai2025p2} & 0.072 & 0.155 & 0.069 & 0.124 & 0.199 & 0.119 & 0.378 \\
      HippoRAG 2 \cite{gutierrez2024hipporag,gutierrez2025hipporag2} & 0.149 & 0.267 & 0.166 & 0.152 & 0.336 & 0.125 & 0.540 \\
      Wikontic \cite{chepurova2025wikontic} & 0.135 & \bf 0.351 & 0.107 & 0.290 & \bf 0.522 & 0.253 & 0.523 \\
      A-RAG \cite{a-rag2026} & 0.057 & 0.047 & 0.102 & 0.129 & 0.164 & 0.124 & 0.331 \\
      TableRAG \cite{yu2025tableragretrievalaugmentedgeneration} & 0.091 & 0.123 & 0.093 & 0.118 & 0.237 & 0.110 & 0.453 \\
      \bf Router with TKG fallback (our) & \bf 0.298 & 0.350 & \bf 0.312 & 0.357 & 0.490 & 0.344 & 0.640 \\
      \bf SKG with TKG fallback (our) & 0.293 & 0.346 & 0.309 & \bf 0.372 & 0.518 & \bf 0.350 & \textbf{0.644} \\
      \bottomrule
    \end{tabular}%
  }
\end{table*}

\begin{table*}
\caption{Main results: comparison between DualGraph and state-of-the-art baselines in terms of token usage.}  
\label{tab:results_usage_sota_full}
  \resizebox{\linewidth}{!}{%
    \begin{tabular}{lS[table-format=9]S[table-format=8] S[table-format=9] S[table-format=6] S[table-format=4] S[table-format=6]}
      \toprule
       & \multicolumn{3}{c}{\textbf{Indexing}} & \multicolumn{3}{c}{\textbf{Querying}} \\
       \cmidrule[0.5pt](lr{0.3em}){2-4} 
       \cmidrule[0.5pt](lr{0.3em}){5-7} 
       & \multicolumn{1}{c}{\textbf{Prompt}} & \multicolumn{1}{c}{\textbf{Completion}} & \multicolumn{1}{c}{\textbf{Total}} & \multicolumn{1}{c}{\textbf{Prompt}} & \multicolumn{1}{c}{\textbf{Completion}} & \multicolumn{1}{c}{\textbf{Total}} \\
      \midrule
      
      LLM-only & 0 & 0 & 0 & 128 & 437 & 565 \\
      VectorRAG & 0 & 0 & 0 & 2215 & 221 & 2436 \\
      Microsoft GraphRAG \cite{edge2025localglobalgraphrag} & 120886386 & 60326338 & 181212724 & 10486 & 634 & 11120 \\
      Microsoft GraphRAG (fast) \cite{edge2025localglobalgraphrag} & 22438336 & 6981861 & 29420197 & 11777 & 459 & 12236 \\
      RAPTOR \cite{DBLP:conf/iclr/SarthiATKGM24} & 4795069 & 819578 & 5614647 & 2078 & 300 & 2378 \\
      LinearRAG \cite{zhuang2025linearraglineargraphretrieval} & 0 & 0 & 0 & 7030 & 511 & 7541 \\
      AriGraph \cite{ijcai2025p2} & 7564910 & 8951382 & 16516292 & 4397 & 700 & 5097 \\
      HippoRAG 2 \cite{gutierrez2024hipporag,gutierrez2025hipporag2} & 10293782 & 5790019 & 16083801 & 21469 & 748 & 22217 \\
      Wikontic\cite{chepurova2025wikontic}  & 364334817 & 55989918 & 420324735 & 133701 & 4926 & 138627 \\
      A-RAG \cite{a-rag2026} & 0 & 0 & 0 & 35712 & 672 & 36384 \\
      TableRAG \cite{yu2025tableragretrievalaugmentedgeneration} & 0 & 0 & 0 & 18526 & 2001 & 20527 \\
      \bf Router with TKG fallback (our) & 7257303 & 8800795 & 16058098 & 4898 & 2780 & 7678 \\
      \bf SKG with TKG fallback (our) & 7257303 & 8800795 & 16058098 & 3813 & 2989 & 6802 \\
      \bottomrule
    \end{tabular}%
  }
\end{table*}

\newpage

\begin{table*}
\caption{Ablation study of DualGraph variants in terms of answer quality and token usage.}  \label{tab:ablation_performance_full}
  \centering
  \resizebox{\linewidth}{!}{%
    \begin{tabular}{lcccccccS[table-format=5]S[table-format=4]S[table-format=5]}
      \toprule
      & \multicolumn{7}{c}{\textbf{Answer Quality }} & \multicolumn{3}{c}{\textbf{Token Usage}} \\
      \cmidrule[0.5pt](lr{0.3em}){2-8}
      \cmidrule[0.5pt](lr{0.3em}){9-11}
       & \multicolumn{3}{c}{\textbf{Factual Correctness}} & \multicolumn{3}{c}{\textbf{List Matching}} & \textbf{LaaJ} & \multicolumn{3}{c}{\textbf{Querying}}\\
       \cmidrule[0.5pt](lr{0.3em}){2-4} 
       \cmidrule[0.5pt](lr{0.3em}){5-7} 
       \cmidrule[0.5pt](lr{0.3em}){9-11} 
       & \multicolumn{1}{c}{\textbf{F1}} & \multicolumn{1}{c}{\textbf{Precision}} & \multicolumn{1}{c}{\textbf{Recall}} & \multicolumn{1}{c}{\textbf{F1}} & \multicolumn{1}{c}{\textbf{Precision}} & \multicolumn{1}{c}{\textbf{Recall}}  & &\multicolumn{1}{c}{\textbf{Prompt}}  & \multicolumn{1}{c}{\textbf{Completions}}  & \multicolumn{1}{c}{\textbf{Total}}  \\
      \midrule
      SKG only              & 0.273 & 0.306 & 0.288 & 0.321 & 0.421 & 0.300 & 0.507 & \bf 2431 & 3040 & 5471 \\
      TKG only              & 0.140 & 0.232 & 0.129 & 0.136 & 0.304 & 0.120 & 0.352 & 2853 & \bf 242 & \bf 3095 \\
      SKG concat TKG        & \bf 0.306 & \bf 0.354 & 0.324 & 0.367 & 0.514 & 0.344 & 0.561 & 5166 & 3020 & 8186 \\
      \bf SKG + TKG\_fb      & 0.293 & 0.346 & 0.309 & \bf 0.372 & \bf 0.518 & \bf 0.350 & 0.523 & 3813 & 2989 & 6802 \\
      Router                & 0.268 & 0.301 & 0.286 & 0.303 & 0.389 & 0.288 & 0.485 & 3720 & 2783 & 6503 \\
      \bf Router + TKG\_fb   & 0.298 & 0.350 & 0.312 & 0.357 & 0.490 & 0.344 & 0.528  & 4898 & 2780 & 7678 \\
      Agentic router        & 0.240 & 0.219 & 0.362 & 0.341 & 0.485 & 0.333 & \textbf{0.764} & 52920 & 2534 & 55454 \\
      \bottomrule
    \end{tabular}%
  }
\end{table*}

\begin{table*}

  \caption{Performance on objective and subjective questions for DualGraph and state-of-the-art baselines.}
  \label{tab:results_performance_objective_sota}
  \centering
  \resizebox{0.9\linewidth}{!}{%
    \begin{tabular}{lrrrrrrr}
      \toprule
       & \multicolumn{3}{c}{\textbf{Factual Correctness}} & \multicolumn{3}{c}{\textbf{List Matching}}  & \textbf{LaaJ} \\
       \cmidrule[0.5pt](lr{0.3em}){2-4} 
       \cmidrule[0.5pt](lr{0.3em}){5-7} 
       & \multicolumn{1}{c}{\textbf{F1}} & \multicolumn{1}{c}{\textbf{Precision}} & \multicolumn{1}{c}{\textbf{Recall}} & \multicolumn{1}{c}{\textbf{F1}} & \multicolumn{1}{c}{\textbf{Precision}} & \multicolumn{1}{c}{\textbf{Recall}} & \\
      \midrule
      {\cellcolor{lightgray} Objective questions (93) }  \\
      LLM only & 0.129 & 0.160 & 0.137 & 0.036 & 0.049 & 0.034 & 0.531 \\
      Vector RAG & 0.115 & 0.200 & 0.106 & 0.158 & 0.326 & 0.142 & 0.451 \\
      Microsoft GraphRAG \cite{edge2025localglobalgraphrag} & 0.178 & 0.283 & 0.174 & 0.243 & 0.438 & 0.223 & 0.573 \\
      Microsoft GraphRAG (fast) \cite{edge2025localglobalgraphrag} & 0.088 & 0.139 & 0.085 & 0.158 & 0.216 & 0.161 & 0.420 \\
      RAPTOR \cite{DBLP:conf/iclr/SarthiATKGM24} & 0.229 & 0.344 & 0.222 & 0.248 & 0.432 & 0.232 & 0.557 \\
      LinearRAG \cite{zhuang2025linearraglineargraphretrieval} & 0.162 & 0.232 & 0.188 & 0.093 & 0.183 & 0.086 & 0.505 \\
      AriGraph \cite{ijcai2025p2} & 0.086 & 0.167 & 0.083 & 0.140 & 0.214 & 0.135 & 0.378 \\
      HippoRAG 2 \cite{gutierrez2024hipporag,gutierrez2025hipporag2} & 0.168 & 0.313 & 0.184 & 0.182 & 0.406 & 0.148 & 0.526 \\
      Wikontic\cite{chepurova2025wikontic} & 0.158 & 0.399 & 0.129 & 0.349 & 0.621 & 0.301 & 0.554 \\
      A-RAG \cite{a-rag2026} & 0.067 & 0.055 & 0.122 & 0.159 & 0.184 & 0.156 & 0.329 \\
      TableRAG \cite{yu2025tableragretrievalaugmentedgeneration} & 0.104 & 0.147 & 0.101 & 0.139 & 0.278 & 0.132 & 0.441 \\
      \bf Router with TKG fallback (our) & 0.354 & 0.412 & 0.370 & 0.441 & 0.580 & 0.427 & \textbf{0.653} \\
      \bf SKG with TKG fallback (our) & 0.337 & 0.395 & 0.356 & 0.434 & 0.573 & 0.416 & 0.643 \\
      
      \midrule
      {\cellcolor{lightgray} Subjective questions (24) }  \\
      LLM only & 0.022 & 0.043 & 0.012 & 0.006 & 0.036 & 0.003 & \textbf{0.671} \\
      Vector RAG & 0.003 & 0.009 & 0.001 & 0.000 & 0.001 & 0.000 & 0.391 \\
      Microsoft GraphRAG \cite{edge2025localglobalgraphrag} & 0.057 & 0.075 & 0.068 & 0.086 & 0.187 & 0.084 & 0.586 \\
      Microsoft GraphRAG (fast) \cite{edge2025localglobalgraphrag} & 0.061 & 0.096 & 0.074 & 0.088 & 0.172 & 0.077 & 0.418 \\
      RAPTOR \cite{DBLP:conf/iclr/SarthiATKGM24} & 0.120 & 0.137 & 0.113 & 0.120 & 0.197 & 0.107 & 0.611 \\
      LinearRAG \cite{zhuang2025linearraglineargraphretrieval} & 0.042 & 0.060 & 0.066 & 0.064 & 0.139 & 0.059 & 0.608 \\
      AriGraph \cite{ijcai2025p2} & 0.020 & 0.109 & 0.013 & 0.075 & 0.154 & 0.069 & 0.378 \\
      HippoRAG 2 \cite{gutierrez2024hipporag,gutierrez2025hipporag2} & 0.075 & 0.090 & 0.096 & 0.064 & 0.129 & 0.056 & 0.596 \\
      Wikontic\cite{chepurova2025wikontic} & 0.048 & 0.173 & 0.027 & 0.118 & 0.231 & 0.110 & 0.403 \\
      A-RAG \cite{a-rag2026} & 0.017 & 0.016 & 0.026 & 0.043 & 0.110 & 0.035 & 0.337 \\
      TableRAG \cite{yu2025tableragretrievalaugmentedgeneration} & 0.039 & 0.036 & 0.062 & 0.053 & 0.115 & 0.045 & 0.497 \\
      \bf Router with TKG fallback (our) & 0.082 & 0.109 & 0.086 & 0.111 & 0.226 & 0.099 & 0.589 \\
      \bf SKG with TKG fallback (our) & 0.124 & 0.154 & 0.129 & 0.188 & 0.355 & 0.157 & 0.649 \\
      \bottomrule
    \end{tabular}%
  }
\end{table*}

\begin{table*}
  \caption{Impact of graph patterns on DualGraph retrieval performance for the Router+TKG fallback and SKG+TKG fallback variants.}
  \label{tab:ablation_patterns_full}
  \resizebox{\linewidth}{!}{%
    \begin{tabular}{ccccccccccccc}
      \toprule
       \multicolumn{4}{c}{\textbf{Pattern}} & \multicolumn{2}{c}{\textbf{\% SKG}} & \multicolumn{3}{c}{\textbf{Factual Correctness}} & \multicolumn{3}{c}{\textbf{List Matching}} & \textbf{LaaJ} \\
       \cmidrule[0.5pt](lr{0.3em}){1-4}
       \cmidrule[0.5pt](lr{0.3em}){5-6}
       \cmidrule[0.5pt](lr{0.3em}){7-9} 
       \cmidrule[0.5pt](lr{0.3em}){10-12} 
       \multicolumn{1}{c}{\textbf{Spec}} & \multicolumn{1}{c}{\textbf{Feature}} & \multicolumn{1}{c}{\textbf{Category}} & \multicolumn{1}{c}{\textbf{Singular Node}} & \multicolumn{1}{c}{\textbf{Router}} & \multicolumn{1}{c}{\textbf{Actual}} & \multicolumn{1}{c}{\textbf{F1}} & \multicolumn{1}{c}{\textbf{Precision}} & \multicolumn{1}{c}{\textbf{Recall}} & \multicolumn{1}{c}{\textbf{F1}} & \multicolumn{1}{c}{\textbf{Precision}} & \multicolumn{1}{c}{\textbf{Recall}} \\
        \midrule
        \multicolumn{4}{l}{{\cellcolor{lightgray} \textbf{DualGraph version}: Router + TKG fallback}}  \\
          Y & Y & Y & Y & 84.4 & 41.9 & \bf 0.298 & \bf 0.350 & \bf 0.312 & 0.357 & 0.490 & 0.344 & \textbf{0.540} \\
          N & Y & Y & Y & 83.1 & 28.9 & 0.225 & 0.300 & 0.232 & 0.274 & 0.439 & 0.256 & 0.470 \\
          Y & N & Y & Y & 83.0 & 40.6 & 0.274 & 0.339 & 0.286 & 0.346 & 0.480 & 0.329 & 0.534 \\
          Y & Y & N & Y & 82.9 & 42.2 & 0.284 & 0.339 & 0.302 & \bf 0.364 & \bf 0.491 & \bf 0.350 & 0.531 \\
          Y & Y & Y & N & 83.5 & 40.2 & 0.266 & 0.324 & 0.282 & 0.339 & 0.475 & 0.325 & 0.527 \\
          Y & N & N & N & 82.6 & 38.5 & 0.271 & 0.331 & 0.288 & 0.349 & 0.484 & 0.334 & 0.532 \\
          N & Y & N & N & 80.9 & 25.4 & 0.228 & 0.303 & 0.236 & 0.266 & 0.422 & 0.251 & 0.473 \\
          N & N & Y & N & 78.3 & 25.8 & 0.206 & 0.283 & 0.211 & 0.243 & 0.399 & 0.226 & 0.454 \\
          N & N & N & Y & 79.2 & 28.0 & 0.230 & 0.296 & 0.237 & 0.265 & 0.407 & 0.253 & 0.468 \\
          N & N & N & N & 76.6 & 24.1 & 0.214 & 0.279 & 0.223 & 0.242 & 0.385 & 0.229 & 0.463 \\
      \midrule
      \multicolumn{4}{l}{{\cellcolor{lightgray} \textbf{DualGraph version}: SKG + TKG fallback}}  \\
          Y & Y & Y & Y & - & 50.2 & 0.293 & 0.346 & 0.309 & 0.372 & \bf 0.518 & 0.350 & 0.534 \\
          N & Y & Y & Y & - & 37.7 & 0.247 & 0.318 & 0.254 & 0.292 & 0.460 & 0.272 & 0.461 \\
          Y & N & Y & Y & - & 48.4 & 0.299 & 0.361 & 0.307 & 0.361 & 0.508 & 0.340 & 0.521 \\
          Y & Y & N & Y & - & 48.8 & \bf 0.317 & \bf 0.364 & \bf 0.331 & \bf 0.380 & 0.512 & \bf 0.360 & 0.538 \\
          Y & Y & Y & N & - & 48.8 & 0.296 & 0.354 & 0.303 & 0.367 & 0.516 & 0.344 & 0.536 \\
          Y & N & N & N & - & 44.7 & 0.301 & 0.351 & 0.314 & 0.362 & 0.493 & 0.345 & \textbf{0.545} \\
          N & Y & N & N & - & 31.3 & 0.246 & 0.316 & 0.254 & 0.290 & 0.446 & 0.272 & 0.478 \\
          N & N & Y & N & - & 34.7 & 0.228 & 0.300 & 0.234 & 0.259 & 0.419 & 0.243 & 0.449 \\
          N & N & N & Y & - & 37.0 & 0.246 & 0.316 & 0.257 & 0.299 & 0.462 & 0.287 & 0.472 \\
          N & N & N & N & - & 29.4 & 0.230 & 0.304 & 0.234 & 0.253 & 0.409 & 0.238 & 0.459 \\
      \bottomrule
    \end{tabular}%
  }
\end{table*}


\begin{thebibliography}{56}
\providecommand{\natexlab}[1]{#1}

\bibitem[{Anokhin et~al.(2025)Anokhin, Semenov, Sorokin, Evseev, Kravchenko,
  Burtsev, and Burnaev}]{ijcai2025p2}
Petr Anokhin, Nikita Semenov, Artyom Sorokin, Dmitry Evseev, Andrey Kravchenko,
  Mikhail Burtsev, and Evgeny Burnaev. 2025.
\newblock \href {https://doi.org/10.24963/ijcai.2025/2} {Arigraph: Learning
  knowledge graph world models with episodic memory for llm agents}.
\newblock In \emph{Proceedings of the Thirty-Fourth International Joint
  Conference on Artificial Intelligence, {IJCAI-25}}, pages 12--20.
  International Joint Conferences on Artificial Intelligence Organization.
\newblock Main Track.

\bibitem[{Berant et~al.(2013)Berant, Chou, Frostig, and
  Liang}]{berant-etal-2013-semantic}
Jonathan Berant, Andrew Chou, Roy Frostig, and Percy Liang. 2013.
\newblock \href {https://aclanthology.org/D13-1160/} {Semantic parsing on
  {F}reebase from question-answer pairs}.
\newblock In \emph{Proceedings of the 2013 Conference on Empirical Methods in
  Natural Language Processing}, pages 1533--1544, Seattle, Washington, USA.
  Association for Computational Linguistics.

\bibitem[{Besrour et~al.(2025)Besrour, He, Schreieder, and
  F{\"a}rber}]{ragenta2025}
Ines Besrour, Jingbo He, Tobias Schreieder, and Michael F{\"a}rber. 2025.
\newblock \href {https://arxiv.org/abs/2506.16988} {{RAGentA}: Multi-agent
  retrieval-augmented generation for attributed question answering}.
\newblock In \emph{SIGIR 2025 LiveRAG Challenge (Workshop)}.

\bibitem[{Chang et~al.(2025)Chang, Jiang, Rakesh, Pan, Yeh, Wang, Hu, Xu,
  Zheng, Das, and Zou}]{chang2025main}
Chia-Yuan Chang, Zhimeng Jiang, Vineeth Rakesh, Menghai Pan, Chin-Chia~Michael
  Yeh, Guanchu Wang, Mingzhi Hu, Zhichao Xu, Yan Zheng, Mahashweta Das, and
  Na~Zou. 2025.
\newblock \href {https://doi.org/10.18653/v1/2025.acl-long.131} {{MAIN}-{RAG}:
  Multi-agent filtering retrieval-augmented generation}.
\newblock In \emph{Proceedings of the 63rd Annual Meeting of the Association
  for Computational Linguistics (ACL 2025)}, pages 2607--2622.

\bibitem[{Chen et~al.(2020)Chen, Zha, Chen, Xiong, Wang, and
  Wang}]{chen-etal-2020-hybridqa}
Wenhu Chen, Hanwen Zha, Zhiyu Chen, Wenhan Xiong, Hong Wang, and William~Yang
  Wang. 2020.
\newblock \href {https://aclanthology.org/2020.findings-emnlp.91/} {{HybridQA}:
  A dataset of multi-hop question answering over tabular and textual data}.
\newblock In \emph{Findings of EMNLP 2020}, pages 1026--1036.

\bibitem[{Chen et~al.(2021)Chen, Chen, Smiley, Shah, Borova, Langdon, Moussa,
  Beane, Huang, Routledge, and Wang}]{chen-etal-2021-finqa}
Zhiyu Chen, Wenhu Chen, Charese Smiley, Sameena Shah, Iana Borova, Dylan
  Langdon, Reema Moussa, Matt Beane, Ting-Hao Huang, Bryan Routledge, and
  William~Yang Wang. 2021.
\newblock \href {https://aclanthology.org/2021.emnlp-main.300/} {{FinQA}: A
  dataset of numerical reasoning over financial data}.
\newblock In \emph{Proceedings of EMNLP 2021}, pages 3697--3711.

\bibitem[{Chen et~al.(2022)Chen, Li, Smiley, Ma, Shah, and
  Wang}]{chen-etal-2022-convfinqa}
Zhiyu Chen, Shiyang Li, Charese Smiley, Zhiqiang Ma, Sameena Shah, and
  William~Yang Wang. 2022.
\newblock \href {https://doi.org/10.18653/v1/2022.emnlp-main.421}
  {{C}onv{F}in{QA}: Exploring the chain of numerical reasoning in
  conversational finance question answering}.
\newblock In \emph{Proceedings of the 2022 Conference on Empirical Methods in
  Natural Language Processing}, pages 6279--6292, Abu Dhabi, United Arab
  Emirates. Association for Computational Linguistics.

\bibitem[{Chen et~al.(2026)Chen, Zheng, and Zhu}]{agenticRAGSurvey2026}
Zihan Chen, Lei Zheng, and Di~Zhu. 2026.
\newblock \href {https://papers.ssrn.com/sol3/papers.cfm?abstract_id=6713979}
  {A survey of agentic graphrag: From retrieval-augmented generation to
  graph-native agents}.
\newblock (6713979).

\bibitem[{Chepurova et~al.(2025)Chepurova, Bulatov, Kuratov, and
  Burtsev}]{chepurova2025wikontic}
Alla Chepurova, Aydar Bulatov, Yuri Kuratov, and Mikhail Burtsev. 2025.
\newblock \href {https://arxiv.org/abs/2512.00590} {Wikontic: Constructing
  wikidata-aligned, ontology-aware knowledge graphs with large language
  models}.
\newblock \emph{Preprint}, arXiv:2512.00590.

\bibitem[{Du et~al.(2026)Du, Xu, Zhu, Wang, Wang, Wang, and Mao}]{a-rag2026}
Mingxuan Du, Benfeng Xu, Chiwei Zhu, Shaohan Wang, Pengyu Wang, Xiaorui Wang,
  and Zhendong Mao. 2026.
\newblock \href {https://arxiv.org/abs/2602.03442} {A-rag: Scaling agentic
  retrieval-augmented generation via hierarchical retrieval interfaces}.
\newblock \emph{arXiv preprint arXiv:2602.03442}.

\bibitem[{Edge et~al.(2025)Edge, Trinh, Cheng, Bradley, Chao, Mody, Truitt,
  Metropolitansky, Ness, and Larson}]{edge2025localglobalgraphrag}
Darren Edge, Ha~Trinh, Newman Cheng, Joshua Bradley, Alex Chao, Apurva Mody,
  Steven Truitt, Dasha Metropolitansky, Robert~Osazuwa Ness, and Jonathan
  Larson. 2025.
\newblock \href {https://arxiv.org/abs/2404.16130} {From local to global: A
  graph rag approach to query-focused summarization}.
\newblock \emph{Preprint}, arXiv:2404.16130.

\bibitem[{Emonet et~al.(2025)Emonet, Bolleman, Duvaud, Mendes~de Farias, and
  Sima}]{emonet2025llm}
Vincent Emonet, Jerven Bolleman, Severine Duvaud, Tarcisio Mendes~de Farias,
  and Ana~Claudia Sima. 2025.
\newblock Llm-based sparql query generation from natural language over
  federated knowledge graphs.
\newblock In \emph{ISWC 2024 Special Session on Harmonising Generative AI and
  Semantic Web Technologies, November 13, 2024, Baltimore, Maryland}, volume
  3953 of \emph{CEUR Workshop Proceedings}.
\newblock CEUR-WS.org, online \url{https://ceur-ws.org/Vol-3953/355.pdf}.

\bibitem[{Friel et~al.(2024)Friel, Belyi, and Sanyal}]{friel-2024-ragbench}
Robert Friel, Masha Belyi, and Atindriyo Sanyal. 2024.
\newblock \href {https://arxiv.org/abs/2407.11005} {{RAGBench}: Explainable
  benchmark for retrieval-augmented generation systems}.
\newblock \emph{arXiv preprint arXiv:2407.11005}.

\bibitem[{Gao et~al.(2023)Gao, Xiong, Gao, Jia, Pan, Bi, Dai, Sun, Wang, Wang
  et~al.}]{gao2023retrieval}
Yunfan Gao, Yun Xiong, Xinyu Gao, Kangxiang Jia, Jinliu Pan, Yuxi Bi, Yixin
  Dai, Jiawei Sun, Haofen Wang, Haofen Wang, and 1 others. 2023.
\newblock Retrieval-augmented generation for large language models: A survey.
\newblock \emph{arXiv preprint arXiv:2312.10997}, 2(1):32.

\bibitem[{Gutiérrez et~al.(2024)Gutiérrez, Shu, Gu, Yasunaga, and
  Su}]{gutierrez2024hipporag}
Bernal~Jiménez Gutiérrez, Yiheng Shu, Yu~Gu, Michihiro Yasunaga, and Yu~Su.
  2024.
\newblock \href {https://openreview.net/forum?id=hkujvAPVsg} {Hipporag:
  Neurobiologically inspired long-term memory for large language models}.
\newblock In \emph{The Thirty-eighth Annual Conference on Neural Information
  Processing Systems}.

\bibitem[{Gutiérrez et~al.(2025)Gutiérrez, Shu, Qi, Zhou, and
  Su}]{gutierrez2025hipporag2}
Bernal~Jiménez Gutiérrez, Yiheng Shu, Weijian Qi, Sizhe Zhou, and Yu~Su.
  2025.
\newblock \href {https://arxiv.org/abs/2502.14802} {From rag to memory:
  Non-parametric continual learning for large language models}.
\newblock \emph{Preprint}, arXiv:2502.14802.

\bibitem[{Ho et~al.(2020)Ho, Nguyen, Sugawara, and Aizawa}]{2WikiMultiHopQA}
Xanh Ho, Anh-Khoa~Duong Nguyen, Saku Sugawara, and Akiko Aizawa. 2020.
\newblock \href {https://arxiv.org/abs/2011.01060} {Constructing a multi-hop qa
  dataset for comprehensive evaluation of reasoning steps}.
\newblock \emph{Preprint}, arXiv:2011.01060.

\bibitem[{Hu et~al.(2024)Hu, Dong, Luo, Han, and Zhang}]{hu-etal-2024-ketqa}
Mengkang Hu, Haoyu Dong, Ping Luo, Shi Han, and Dongmei Zhang. 2024.
\newblock \href {https://arxiv.org/abs/2405.08099} {{KET-QA}: A dataset for
  knowledge enhanced table question answering}.
\newblock \emph{arXiv preprint arXiv:2405.08099}.

\bibitem[{Hu et~al.(2025)Hu, Lei, Zhang, Pan, Ling, and
  Zhao}]{hu2025graggraphretrievalaugmentedgeneration}
Yuntong Hu, Zhihan Lei, Zheng Zhang, Bo~Pan, Chen Ling, and Liang Zhao. 2025.
\newblock \href {https://arxiv.org/abs/2405.16506} {Grag: Graph
  retrieval-augmented generation}.
\newblock \emph{Preprint}, arXiv:2405.16506.

\bibitem[{Joshi et~al.(2017)Joshi, Choi, Weld, and
  Zettlemoyer}]{joshi-etal-2017-triviaqa}
Mandar Joshi, Eunsol Choi, Daniel Weld, and Luke Zettlemoyer. 2017.
\newblock \href {https://doi.org/10.18653/v1/P17-1147} {{T}rivia{QA}: A large
  scale distantly supervised challenge dataset for reading comprehension}.
\newblock In \emph{Proceedings of the 55th Annual Meeting of the Association
  for Computational Linguistics (Volume 1: Long Papers)}, pages 1601--1611,
  Vancouver, Canada. Association for Computational Linguistics.

\bibitem[{Kong et~al.(2024)Kong, Zhang, Shen, Srinivasan, Lei, Faloutsos,
  Rangwala, and Karypis}]{kong2024opentab}
Kezhi Kong, Jiani Zhang, Zhengyuan Shen, Balasubramaniam Srinivasan, Chuan Lei,
  Christos Faloutsos, Huzefa Rangwala, and George Karypis. 2024.
\newblock Opentab: Advancing large language models as open-domain table
  reasoners.
\newblock \emph{arXiv preprint arXiv:2402.14361}.
\newblock ICLR 2024, Code:
  \url{https://github.com/amazon-science/llm-open-domain-table-reasoner}.

\bibitem[{Kwiatkowski et~al.(2019)Kwiatkowski, Palomaki, Redfield, Collins,
  Parikh, Alberti, Epstein, Polosukhin, Devlin, Lee, Toutanova, Jones, Kelcey,
  Chang, Dai, Uszkoreit, Le, and Petrov}]{kwiatkowski-etal-2019-natural}
Tom Kwiatkowski, Jennimaria Palomaki, Olivia Redfield, Michael Collins, Ankur
  Parikh, Chris Alberti, Danielle Epstein, Illia Polosukhin, Jacob Devlin,
  Kenton Lee, Kristina Toutanova, Llion Jones, Matthew Kelcey, Ming-Wei Chang,
  Andrew~M. Dai, Jakob Uszkoreit, Quoc Le, and Slav Petrov. 2019.
\newblock \href {https://doi.org/10.1162/tacl_a_00276} {Natural questions: A
  benchmark for question answering research}.
\newblock \emph{Transactions of the Association for Computational Linguistics},
  7:452--466.

\bibitem[{Lewis et~al.(2020)Lewis, Perez, Piktus, Petroni, Karpukhin, Goyal,
  K\"{u}ttler, Lewis, Yih, Rockt\"{a}schel, Riedel, and Kiela}]{2020_RAG_paper}
Patrick Lewis, Ethan Perez, Aleksandra Piktus, Fabio Petroni, Vladimir
  Karpukhin, Naman Goyal, Heinrich K\"{u}ttler, Mike Lewis, Wen-tau Yih, Tim
  Rockt\"{a}schel, Sebastian Riedel, and Douwe Kiela. 2020.
\newblock \href
  {https://proceedings.neurips.cc/paper_files/paper/2020/file/6b493230205f780e1bc26945df7481e5-Paper.pdf}
  {Retrieval-augmented generation for knowledge-intensive nlp tasks}.
\newblock In \emph{Advances in Neural Information Processing Systems},
  volume~33, pages 9459--9474. Curran Associates, Inc.

\bibitem[{Li et~al.(2025{\natexlab{a}})Li, Fang, Shi, Khan, Wang, Wang,
  Zhangxin-hw, and Cui}]{li2025cot}
Feiyang Li, Peng Fang, Zhan Shi, Arijit Khan, Fang Wang, Weihao Wang,
  Zhangxin-hw, and Yongjian Cui. 2025{\natexlab{a}}.
\newblock \href {https://doi.org/10.18653/v1/2025.findings-emnlp.168}
  {{CoT}-{RAG}: Integrating chain of thought and retrieval-augmented generation
  to enhance reasoning in large language models}.
\newblock In \emph{Findings of the Association for Computational Linguistics:
  EMNLP 2025}, pages 3119--3171.

\bibitem[{Li et~al.(2025{\natexlab{b}})Li, Zhang, Yang, Huang, Wu, Luo, Bei,
  Zou, Luo, Zhao et~al.}]{li2025survey}
Yangning Li, Weizhi Zhang, Yuyao Yang, Wei-Chieh Huang, Yaozu Wu, Junyu Luo,
  Yuanchen Bei, Henry~Peng Zou, Xiao Luo, Yusheng Zhao, and 1 others.
  2025{\natexlab{b}}.
\newblock A survey of rag-reasoning systems in large language models.
\newblock In \emph{Findings of the Association for Computational Linguistics:
  EMNLP 2025}, pages 12120--12145.

\bibitem[{Liu et~al.(2025)Liu, Liu, Yao, Liu, Meng, Wang, and
  Ma}]{liu2025hmrag}
Pei Liu, Xin Liu, Ruoyu Yao, Junming Liu, Siyuan Meng, Ding Wang, and Jun Ma.
  2025.
\newblock \href {https://arxiv.org/abs/2504.12330} {{HM}-{RAG}: Hierarchical
  multi-agent multimodal retrieval augmented generation}.
\newblock \emph{arXiv preprint arXiv:2504.12330}.

\bibitem[{Maragheh et~al.(2025)Maragheh, Vadla, Gupta, Zhao, Inan, Yao, Xu,
  Kanumala, Cho, and Kumar}]{arag2025}
Reza~Yousefi Maragheh, Pratheek Vadla, Priyank Gupta, Kai Zhao, Aysenur Inan,
  Kehui Yao, Jianpeng Xu, Praveen Kanumala, Jason Cho, and Sushant Kumar. 2025.
\newblock \href {https://arxiv.org/abs/2506.21931} {{ARAG}: Agentic retrieval
  augmented generation for personalized recommendation}.
\newblock In \emph{Proceedings of the 48th ACM SIGIR Conference (SIGIR 2025)}.

\bibitem[{Mo et~al.(2025)Mo, Yu, Kazdan, Mpala, Yu, Cundy, Kanatsoulis, and
  Koyejo}]{mo2025kggen}
Belinda Mo, Kyssen Yu, Joshua Kazdan, Proud Mpala, Lisa Yu, Chris Cundy,
  Charilaos~I. Kanatsoulis, and Sanmi Koyejo. 2025.
\newblock Kggen: Extracting knowledge graphs from plain text with language
  models.
\newblock \emph{CoRR}, abs/2502.09956.

\bibitem[{Nguyen et~al.(2025)Nguyen, Chin, and Tai}]{nguyen2025marag}
Thang Nguyen, Peter Chin, and Yu-Wing Tai. 2025.
\newblock \href {https://arxiv.org/abs/2505.20096} {{MA}-{RAG}: Multi-agent
  retrieval-augmented generation via collaborative chain-of-thought reasoning}.
\newblock \emph{arXiv preprint arXiv:2505.20096}.

\bibitem[{Nizar et~al.(2025)Nizar, Lumer, Gulati, Basavaraju, and
  Subbiah}]{nizar2025agentag}
Faheem Nizar, Elias Lumer, Anmol Gulati, Pradeep Basavaraju, and Vamse~Kumar
  Subbiah. 2025.
\newblock \href {https://arxiv.org/abs/2511.18194} {{Agent}-as-a-{Graph}:
  Knowledge graph-based tool and agent retrieval for llm multi-agent systems}.
\newblock \emph{arXiv preprint arXiv:2511.18194}.

\bibitem[{OpenAI(2025)}]{openai2025gptoss120bgptoss20bmodel}
OpenAI. 2025.
\newblock \href {https://arxiv.org/abs/2508.10925} {gpt-oss-120b \& gpt-oss-20b
  model card}.
\newblock \emph{Preprint}, arXiv:2508.10925.

\bibitem[{Pasupat and Liang(2015)}]{pasupat-liang-2015-compositional}
Panupong Pasupat and Percy Liang. 2015.
\newblock \href {https://aclanthology.org/P15-1142/} {Compositional semantic
  parsing on semi-structured tables}.
\newblock In \emph{Proceedings of ACL 2015}, pages 1470--1480.

\bibitem[{Peng et~al.(2025)Peng, Zhu, Liu, Bo, Shi, Hong, Zhang, and
  Tang}]{peng2025graphragSurvey}
Boci Peng, Yun Zhu, Yongchao Liu, Xiaohe Bo, Haizhou Shi, Chuntao Hong, Yan
  Zhang, and Siliang Tang. 2025.
\newblock \href {https://doi.org/10.1145/3777378} {Graph retrieval-augmented
  generation: A survey}.
\newblock \emph{ACM Trans. Inf. Syst.}, 44(2).

\bibitem[{Pydantic(2023)}]{pydantic-ai:online}
Pydantic. 2023.
\newblock \href {https://github.com/pydantic/pydantic-ai}
  {pydantic/pydantic-ai: Genai agent framework, the pydantic way}.
\newblock [Online; accessed 2026-01-30].

\bibitem[{Roy et~al.(2025)Roy, Hinze, Schlotthauer, Naderi, Hangya, Foltyn,
  Hahn, and K{\"{u}}ch}]{ragonite}
Rishiraj~Saha Roy, Chris Hinze, Joel Schlotthauer, Farzad Naderi, Viktor
  Hangya, Andreas Foltyn, Luzian Hahn, and Fabian K{\"{u}}ch. 2025.
\newblock \href {https://doi.org/10.18420/BTW2025-43} {{RAGONITE:} iterative
  retrieval on induced databases and verbalized {RDF} for conversational {QA}
  over kgs with {RAG}}.
\newblock In \emph{Datenbanksysteme f{\"{u}}r Business, Technologie und Web
  {(BTW} 2025), 21. Fachtagung des GI-Fachbereichs ,,Datenbanken und
  Informationssysteme" (DBIS), 03.-07, M{\"{a}}rz 2025, Bamberg, Germany,
  Proceedings}, volume {P-361} of \emph{{LNI}}, pages 787--794. Gesellschaft
  f{\"{u}}r Informatik e.V.

\bibitem[{Roy et~al.(2024)Roy, Hinze, Schlotthauer, Naderi, Hangya, Foltyn,
  Hahn, and Kuech}]{roy2024ragoniteiterativeretrievalinduced}
Rishiraj~Saha Roy, Chris Hinze, Joel Schlotthauer, Farzad Naderi, Viktor
  Hangya, Andreas Foltyn, Luzian Hahn, and Fabian Kuech. 2024.
\newblock \href {https://arxiv.org/abs/2412.17690} {Ragonite: Iterative
  retrieval on induced databases and verbalized rdf for conversational qa over
  kgs with rag}.
\newblock \emph{Preprint}, arXiv:2412.17690.

\bibitem[{Sarthi et~al.(2024)Sarthi, Abdullah, Tuli, Khanna, Goldie, and
  Manning}]{DBLP:conf/iclr/SarthiATKGM24}
Parth Sarthi, Salman Abdullah, Aditi Tuli, Shubh Khanna, Anna Goldie, and
  Christopher~D. Manning. 2024.
\newblock \href {https://openreview.net/forum?id=GN921JHCRw} {{RAPTOR:}
  recursive abstractive processing for tree-organized retrieval}.
\newblock In \emph{The Twelfth International Conference on Learning
  Representations, {ICLR} 2024, Vienna, Austria, May 7-11, 2024}.
  OpenReview.net.

\bibitem[{Singh et~al.(2025)Singh, Ehtesham, Kumar, and
  Khoei}]{singh2025agentic}
Aditi Singh, Abul Ehtesham, Saket Kumar, and Tala~Talaei Khoei. 2025.
\newblock Agentic retrieval-augmented generation: A survey on agentic rag.
\newblock \emph{arXiv preprint arXiv:2501.09136}.

\bibitem[{Smeros et~al.(2025)Smeros, Emonet, Wang, Sima, and
  de~Farias}]{smeros2025sparqlllmrealtimesparqlquery}
Panayiotis Smeros, Vincent Emonet, Ruijie Wang, Ana-Claudia Sima, and
  Tarcisio~Mendes de~Farias. 2025.
\newblock \href {https://arxiv.org/abs/2512.14277} {Sparql-llm: Real-time
  sparql query generation from natural language questions}.
\newblock \emph{Preprint}, arXiv:2512.14277.

\bibitem[{Strich et~al.(2026)Strich, Isgorur, Trescher, Biemann, and
  Semmann}]{strich-etal-2026-t2}
Jan Strich, Enes~Kutay Isgorur, Maximilian Trescher, Chris Biemann, and Martin
  Semmann. 2026.
\newblock \href {https://doi.org/10.18653/v1/2026.eacl-long.8}
  {T$^2$-{RAGB}ench: Text-and-table benchmark for evaluating
  retrieval-augmented generation}.
\newblock In \emph{Proceedings of the 19th Conference of the {E}uropean Chapter
  of the {A}ssociation for {C}omputational {L}inguistics (Volume 1: Long
  Papers)}, pages 165--191, Rabat, Morocco. Association for Computational
  Linguistics.

\bibitem[{Talmor and Berant(2018)}]{talmor-berant-2018-web}
Alon Talmor and Jonathan Berant. 2018.
\newblock \href {https://doi.org/10.18653/v1/N18-1059} {The web as a
  knowledge-base for answering complex questions}.
\newblock In \emph{Proceedings of the 2018 Conference of the North {A}merican
  Chapter of the Association for Computational Linguistics: Human Language
  Technologies, Volume 1 (Long Papers)}, pages 641--651, New Orleans,
  Louisiana. Association for Computational Linguistics.

\bibitem[{Trivedi et~al.(2022)Trivedi, Balasubramanian, Khot, and
  Sabharwal}]{trivedi-etal-2022-musique}
Harsh Trivedi, Niranjan Balasubramanian, Tushar Khot, and Ashish Sabharwal.
  2022.
\newblock \href {https://aclanthology.org/2022.tacl-1.31/} {{MuSiQue}: Multihop
  questions via single-hop question composition}.
\newblock \emph{TACL}, 10:539--554.

\bibitem[{Tuora et~al.(2026)Tuora, Galiński, Godziszewski, Karpowicz,
  Czyżnikiewicz, Kozakiewicz, and
  Ziętkiewicz}]{tuora2026unweavingknotsgraphrag}
Ryszard Tuora, Mateusz Galiński, Michał Godziszewski, Michał Karpowicz,
  Mateusz Czyżnikiewicz, Adam Kozakiewicz, and Tomasz Ziętkiewicz. 2026.
\newblock \href {https://arxiv.org/abs/2603.29875} {Unweaving the knots of
  graphrag -- turns out vectorrag is almost enough}.
\newblock \emph{Preprint}, arXiv:2603.29875.

\bibitem[{VibrantLabs(2024)}]{ragas2024}
VibrantLabs. 2024.
\newblock Ragas: Supercharge your llm application evaluations.
\newblock \url{https://github.com/vibrantlabsai/ragas}.

\bibitem[{Yang et~al.(2018)Yang, Qi, Zhang, Bengio, Cohen, Salakhutdinov, and
  Manning}]{yang-etal-2018-hotpotqa}
Zhilin Yang, Peng Qi, Saizheng Zhang, Yoshua Bengio, William Cohen, Ruslan
  Salakhutdinov, and Christopher~D. Manning. 2018.
\newblock \href {https://aclanthology.org/D18-1259/} {{HotpotQA}: A dataset for
  diverse, explainable multi-hop question answering}.
\newblock In \emph{Proceedings of EMNLP 2018}, pages 2369--2380.

\bibitem[{Yu et~al.(2018)Yu, Zhang, Yang, Yasunaga, Wang, Li, Ma, Li, Yao,
  Roman, Zhang, and Radev}]{yu-etal-2018-spider}
Tao Yu, Rui Zhang, Kai Yang, Michihiro Yasunaga, Dongxu Wang, Zifan Li, James
  Ma, Irene Li, Qingning Yao, Shanelle Roman, Zilin Zhang, and Dragomir Radev.
  2018.
\newblock \href {https://doi.org/10.18653/v1/D18-1425} {{S}pider: A large-scale
  human-labeled dataset for complex and cross-domain semantic parsing and
  text-to-{SQL} task}.
\newblock In \emph{Proceedings of the 2018 Conference on Empirical Methods in
  Natural Language Processing}, pages 3911--3921, Brussels, Belgium.
  Association for Computational Linguistics.

\bibitem[{Yu et~al.(2025{\natexlab{a}})Yu, Jian, and
  Chen}]{yu2025tableragretrievalaugmentedgeneration}
Xiaohan Yu, Pu~Jian, and Chong Chen. 2025{\natexlab{a}}.
\newblock \href {https://arxiv.org/abs/2506.10380} {Tablerag: A retrieval
  augmented generation framework for heterogeneous document reasoning}.
\newblock \emph{Preprint}, arXiv:2506.10380.

\bibitem[{Yu et~al.(2025{\natexlab{b}})Yu, Jian, and
  Chen}]{yu-etal-2025-tablerag}
Xiaohan Yu, Pu~Jian, and Chong Chen. 2025{\natexlab{b}}.
\newblock \href {https://doi.org/10.18653/v1/2025.emnlp-main.710}
  {{T}able{RAG}: A retrieval augmented generation framework for heterogeneous
  document reasoning}.
\newblock In \emph{Proceedings of the 2025 Conference on Empirical Methods in
  Natural Language Processing}, pages 14063--14082, Suzhou, China. Association
  for Computational Linguistics.

\bibitem[{Yu et~al.(2025{\natexlab{c}})Yu, Jian, and Chen}]{yu2025tablerag}
Xiaohan Yu, Pu~Jian, and Chong Chen. 2025{\natexlab{c}}.
\newblock \href {https://arxiv.org/abs/2506.10380} {Tablerag: A retrieval
  augmented generation framework for heterogeneous document reasoning}.
\newblock In \emph{Proceedings of EMNLP 2025}, pages 14063--14082.
\newblock Code: \url{https://github.com/yxh-y/TableRAG}.

\bibitem[{Yu et~al.(2025{\natexlab{d}})Yu, Yang, and Chen}]{cogplanner2025}
Xiaohan Yu, Zhihan Yang, and Chong Chen. 2025{\natexlab{d}}.
\newblock \href {https://arxiv.org/abs/2501.15470} {{CogPlanner}: Unveiling the
  potential of agentic multimodal retrieval augmented generation with
  planning}.
\newblock In \emph{Proceedings of the 48th International ACM SIGIR Conference
  on Research and Development in Information Retrieval (SIGIR 2025)}.

\bibitem[{Zhang et~al.(2025)Zhang, Li, Long, Zhang, Lin, Yang, Xie, Yang, Liu,
  Lin, Huang, and Zhou}]{qwen3embedding}
Yanzhao Zhang, Mingxin Li, Dingkun Long, Xin Zhang, Huan Lin, Baosong Yang,
  Pengjun Xie, An~Yang, Dayiheng Liu, Junyang Lin, Fei Huang, and Jingren Zhou.
  2025.
\newblock Qwen3 embedding: Advancing text embedding and reranking through
  foundation models.
\newblock \emph{arXiv preprint arXiv:2506.05176}.

\bibitem[{Zhong et~al.(2017)Zhong, Xiong, and
  Socher}]{zhong2017seq2sqlgeneratingstructuredqueries}
Victor Zhong, Caiming Xiong, and Richard Socher. 2017.
\newblock \href {https://arxiv.org/abs/1709.00103} {Seq2sql: Generating
  structured queries from natural language using reinforcement learning}.
\newblock \emph{Preprint}, arXiv:1709.00103.

\bibitem[{Zhou et~al.(2024)Zhou, Huang, Long, Xu, Zhu, Cao, Yang, and
  Zhao}]{zhou2024mitigating}
Hongli Zhou, Hui Huang, Yunfei Long, Bing Xu, Conghui Zhu, Hailong Cao, Muyun
  Yang, and Tiejun Zhao. 2024.
\newblock Mitigating the bias of large language model evaluation.
\newblock In \emph{Proceedings of the 23rd Chinese National Conference on
  Computational Linguistics (Volume 1: Main Conference)}, pages 1310--1319.

\bibitem[{Zhu et~al.(2021)Zhu, Lei, Huang, Wang, Zhang, Lv, Feng, and
  Chua}]{zhu-etal-2021-tat}
Fengbin Zhu, Wenqiang Lei, Youcheng Huang, Chao Wang, Shuo Zhang, Jiancheng Lv,
  Fuli Feng, and Tat-Seng Chua. 2021.
\newblock \href {https://aclanthology.org/2021.acl-long.254/} {{TAT-QA}: A
  question answering benchmark on a hybrid of tabular and textual content in
  finance}.
\newblock In \emph{Proceedings of ACL-IJCNLP 2021}, pages 3277--3287.

\bibitem[{Zhuang et~al.(2025)Zhuang, Chen, Xiao, Zhou, Zhang, Chen, Zhang, and
  Huang}]{zhuang2025linearraglineargraphretrieval}
Luyao Zhuang, Shengyuan Chen, Yilin Xiao, Huachi Zhou, Yujing Zhang, Hao Chen,
  Qinggang Zhang, and Xiao Huang. 2025.
\newblock \href {https://arxiv.org/abs/2510.10114} {Linearrag: Linear graph
  retrieval augmented generation on large-scale corpora}.
\newblock \emph{Preprint}, arXiv:2510.10114.

\bibitem[{Zou et~al.(2025)Zou, Fu, Chen, He, Li, Zhu, Han, and He}]{zou2025rag}
Jiaru Zou, Dongqi Fu, Sirui Chen, Xinrui He, Zihao Li, Yada Zhu, Jiawei Han,
  and Jingrui He. 2025.
\newblock \href {https://arxiv.org/abs/2504.01346} {Rag over tables:
  Hierarchical memory index, multi-stage retrieval, and benchmarking}.
\newblock \emph{arXiv preprint arXiv:2504.01346}.
\newblock Code: \url{https://github.com/jiaruzouu/T-RAG}.

\end{thebibliography}
\end{document}